\documentclass[10pt,twocolumn,letterpaper]{article}
\usepackage{cvpr}
\usepackage{times}
\usepackage{epsfig}
\usepackage{graphicx}
\usepackage{amsmath}
\usepackage{amssymb}
\usepackage{makecell}
\usepackage{xcolor}
\definecolor{medium-blue}{rgb}{0,0,1}
\usepackage{hyperref}
\hypersetup{colorlinks, urlcolor={medium-blue}}

\usepackage{array,multirow,graphicx}
\usepackage{float}
\usepackage[numbers,sort]{natbib}
\usepackage{mathtools}
\DeclareMathOperator*{\argmax}{arg\,max}
\DeclareMathOperator*{\argmin}{arg\,min}
\usepackage{xcolor}
\usepackage{bbm}
\usepackage{arydshln}
\usepackage[linesnumbered,ruled,vlined]{algorithm2e}
\SetKwInput{KwInput}{Input}                
\SetKwInput{KwOutput}{Output}              

\usepackage{enumitem} 
\newenvironment{packed_description}{
\begin{description}[leftmargin=0pt]
  \setlength{\itemsep}{1.5pt}
  \setlength{\parskip}{0pt}
  \setlength{\parsep}{0pt}
}{\end{description}}

\DeclarePairedDelimiter\abs{\lvert}{\rvert}%
\DeclarePairedDelimiter\norm{\lVert}{\rVert}%
\DeclareMathOperator*{\dettime}{time}
\DeclareMathOperator*{\cam}{cam}
\DeclareMathOperator*{\iou}{IoU}
\DeclareMathOperator*{\dist}{dist}
\DeclareMathOperator*{\visible}{visible}
\DeclareMathOperator*{\R}{\mathbb{R}}
\DeclareMathOperator*{\for}{\forall}
\usepackage{tabularx}
\makeatletter
\newcommand{\removelatexerror}{\let\@latex@error\@gobble}
\makeatother
\def\myparagraph#1{\vspace*{3pt}\noindent{\bf #1~~}}

\newcommand{\STAB}[1]{\begin{tabular}{@{}c@{}}#1\end{tabular}}

\makeatletter
\@namedef{ver@everyshi.sty}{}
\makeatother
\usepackage[disable]{todonotes}

\begin{document}

\title{LMGP: Lifted Multicut Meets Geometry Projections for Multi-Camera Multi-Object Tracking}
\author{\vspace{-0.15in} Duy M. H. Nguyen\textsuperscript{\rm 1,4},\ \ 
Roberto Henschel\textsuperscript{\rm 2},\ \ Bodo Rosenhahn\textsuperscript{\rm 2},\ \  Daniel Sonntag\textsuperscript{\rm 3,4},\ \ Paul Swoboda \textsuperscript{\rm 1} \\\\
\textsuperscript{\rm 1} \small{Max Planck Institute for Informatics, Saarland Informatics Campus},
\textsuperscript{\rm 2} \small{Institute for Information Processing, Leibniz University Hannover} \\
\textsuperscript{\rm 3} {\small Oldenburg University}
\textsuperscript{\rm 4} {\small German Research Center for Artificial Intelligence, Saarbrücken}
}

\maketitle
\begin{abstract}
Multi-Camera Multi-Object Tracking is currently drawing attention in the computer vision field due to its superior performance in real-world applications such as video surveillance with crowded scenes or in wide spaces. In this work, we propose a mathematically elegant multi-camera multiple object tracking approach based on a spatial-temporal lifted multicut formulation. Our model utilizes state-of-the-art tracklets produced by single-camera trackers as proposals. As these tracklets may contain ID-Switch errors, we refine them through a novel pre-clustering obtained from 3D geometry projections.  As a result, we derive a better tracking graph without ID switches and more precise affinity costs for the data association phase. Tracklets are then matched to multi-camera trajectories by solving a global lifted multicut formulation that incorporates short and long-range temporal interactions on tracklets located in the same camera as well as inter-camera ones. Experimental results on the WildTrack dataset yield near-perfect performance, outperforming state-of-the-art trackers on Campus while being on par on the PETS-09 dataset. We will release our implementations at this link  \href{https://github.com/nhmduy/LMGP}{https://github.com/nhmduy/LMGP}.

\end{abstract}

\section{Introduction}
\begin{figure}[!hbt]
\centering
\includegraphics[width=0.48\textwidth]{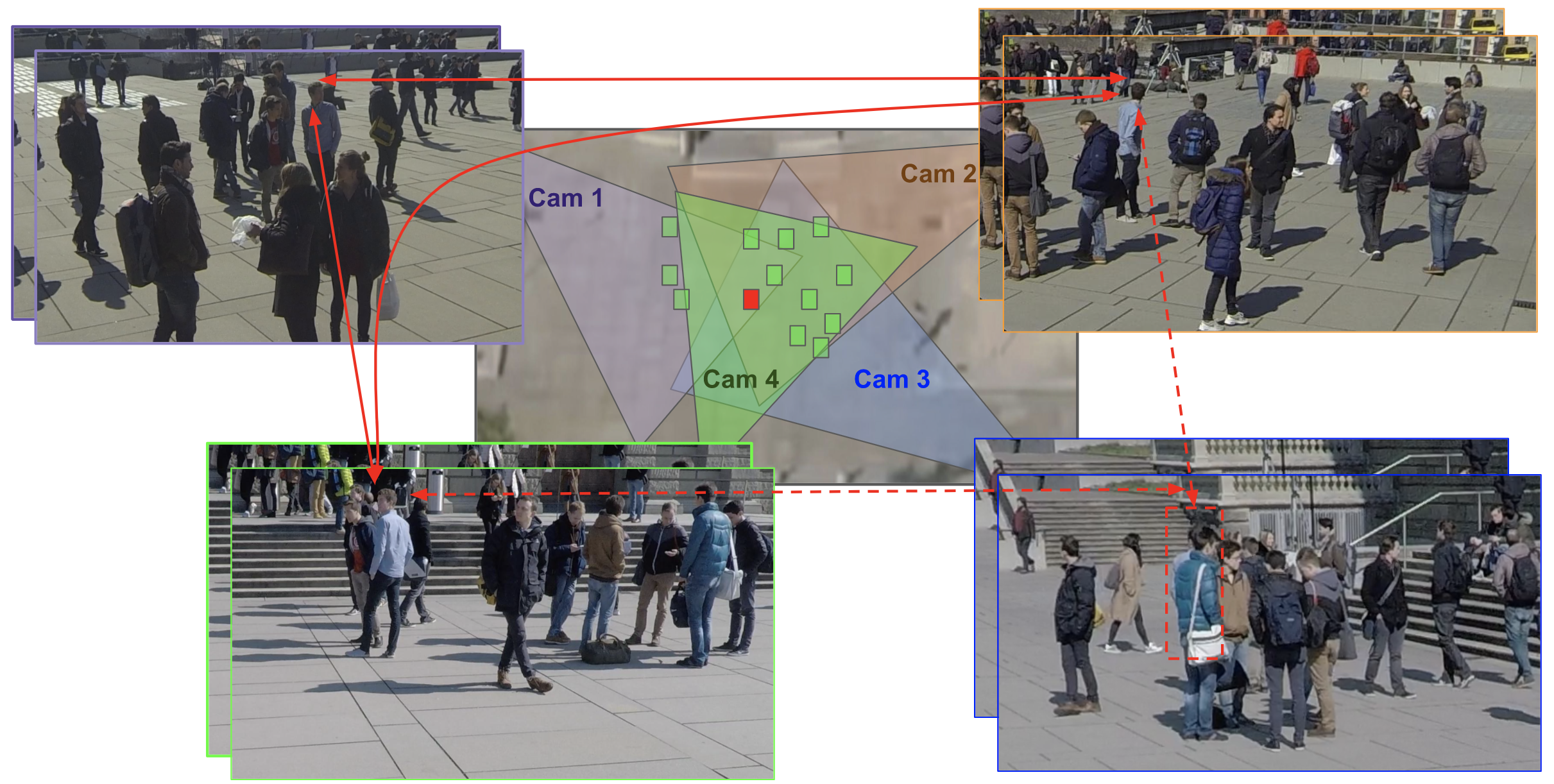}
\caption{
Multi-camera tracking with four overlapping cameras. A target object (red rectangle) is occluded at Cam 3 but is still observed at Cam 1, Cam 2, and Cam 4. Taking this correspondence into account (red arrow), we can recover a missing bounding box at Cam 3 (red dashed arrow).}
\label{fig:teaser_crop}
\vspace{-0.2in}
\end{figure}
Multiple object tracking (MOT), i.e., \ extracting motions of objects moving through a scene, is a fundamental primitive for high-level understanding information in videos.
The most common approach to MOT is the tracking-by-assignment paradigm, in which first detection boxes are computed for the objects of interest in each timeframe, and second, a data association is performed by linking detections of the same objects to each other.
In the most popular setting, a single camera faces a scene and the data association links detections in different timeframes to each other ~\citep{bewley2016simple, wojke2017simple, chu2019famnet,bergmann2019tracking}.
However, even though a large body of research has been devoted to MOT with a single camera, large and crowded scenes still cannot be tracked faithfully, and errors occur mainly in the data association step.
These errors most often are caused by partial visibility (or even occlusion) and indistinguishability of objects.

One possibility of improving performance has been to use multiple cameras facing the same scene but from different angles (Figure \ref{fig:teaser_crop}).
In this setting, partial visibility and indistinguishability are less severe since an object may be occluded in a single camera but may still be fully observed by another camera. Leveraging this property, recent papers have pursued two principal approaches: \textit{single view-based} and \textit{centralized representation} methods. In the first strategy \cite{xu2017cross,he2020multi,lan2020semi,quach2021dyglip}, a two-step procedure is followed: 1) generating local tracklets of all the targets within each camera; 2) matching local tracklets that belong to the same target across cameras through computing affinity costs and using a global optimization framework. While this framework brings benefits via a reduced hypothesis space and allows to design motion-based features, its main drawbacks lie in ID-Switch errors contained inside local tracklets, i.e., detections of distinct objects are grouped into the same trajectory (Figure \ref{fig:split-tracklet}a). As a result, these errors will propagate throughout the tracking graph,
affecting the total performance. \textit{The centralized representation} approach \cite{zhang2019wide,you2020real} on the other hand, is not plagued by such obstacles since each node in the tracking graph is an occupancy map (not a tracklet), which is estimated from all detections at each timeframe. Unfortunately, the cost of the data association step is increased due to a huge state space of variables and integrating advances from single-camera methods is more complicated.


In this work, we propose a method that follows the \textit{single view-based approach} but integrates concepts from the centralized representation paradigm. Our motivation for this design choice is to harness the great progress made in single-camera tracking while at the same time effectively addressing limitations encountered in prior studies such as ID-Switch errors by exploiting \textit{centralized representation} ideas through our novel pre-clustering step.
Specifically, the corresponding images from the pre-clustering step (i.e.\ our occupancy map) allow us to break up initial tracklets generated by single-camera trackers at ID-Switch errors and establish precise affinity costs for both temporal and spatial affinities (Subsection \ref{sub:ablation-study}). On top of that, a novel spatial-temporal optimization model for the data association is employed, which takes into account both short- and long-range temporal interactions of objects detected by a single camera as well as spatial interactions between cameras in a single framework (Table \ref{tab:joint-spatial-temporal-optimize}). The experimental findings show that, given the right conditions, with a multiple-camera environment and precise boundary detections, our method leads to a nearly optimal solution for multiple object tracking using multiple cameras (Table \ref{tab:multi-wildtrack}).

\myparagraph{Contributions} Our main contributions can be summarized as follows. First, we introduce a new pre-clustering algorithm driven by 3D geometry projections to group detections at each timestep across cameras. This effectively eliminates tracklet errors from single-camera trackers and provides highly accurate affinity costs for the data association step.
Second, we propose a novel spatial-temporal lifted multicut formulation for the multi-camera setting, jointly optimizing both intra- and inter-camera as well as short- and long-range interaction in a single global formulation.
Finally, we obtain nearly perfect performance on the large-scale WILDTRACK~\cite{chavdarova2018wildtrack} dataset, outperform state-of-the-art on Campus~\cite{xu2016multi} and are on par with the PETS-09~\cite{ellis2009pets2009} dataset. 
\section{Related Work}
There has been a large body of research on single-camera MOT.
These methods focus on the data association step, for which the (lifted) multicut problem~\cite{tang2015subgraph,tang2016multi,tang2017multiple}, the lifted disjoint paths problem~\cite{hornakova2020lifted,hornakova2021making}, maximum clique~\cite{zamir2012gmcp,dehghan2015gmmcp},
multigraph-matching~\cite{hu2019dual}, and
binary quadratic optimization~\cite{Henschel_2018_CVPR_Workshops,henschel2016tracking,henschel2020simultaneous,von2018recovering} was used.
Another area is building end-to-end differentiable frameworks for both detector and data association~\citep{wojke2017simple,chu2019famnet,bergmann2019tracking,zhou2020tracking,zhang2021fairmot}.
For an exhaustive survey of MOT we refer to \cite{dendorfer2021motchallenge}.

Multi-camera MOT has recently received increasing attention. The related work can be categorized into two different approaches: 

\myparagraph{Single View-Based Methods} 
\cite{xu2016multi} propose a Hierarchical Composition of Tracklet (HCT) framework to match local tracklets by utilizing multiple cues of objects such as appearances and their 3D positions. 
In \cite{xu2017cross} the matching problem is solved using a Bayesian formulation with a Spatio-Temporal Parsing (STP)-based tracking graph to prune matching candidates by exploiting semantic attribute targets.
Similarly,~\cite{wen2017multi} formulate a dense sub-hypergraph search (SVTH) on the space-time-view graph using a sampling-based approach.
Recent approaches include a semi-online Multi-Label Markov Random Field (MLMRF) method~\cite{lan2020semi}, where the ensuing optimization problem over single detections is solved through alpha-expansion~\cite{boykov2001fast} and a non-negative matrix factorization approach (TRACTA) for grouping tracklets across cameras~\cite{he2020multi}.
In another direction, DyGLIP~\cite{quach2021dyglip} formulates the data association problem for multi-camera as a link prediction on a graph whose nodes are tracklets. 
While these methods have demonstrated promising performance in some datasets, they are affected by ID-switch errors in the tracklet proposal generation, especially in cluttered or crowded scenes such as~\cite{chavdarova2018wildtrack}.

\myparagraph{Centralized Representation-Based Methods} To estimate the occupancy map (2D) or occupancy volume (3D), occlusion relationships among different detections have been explicitly modeled. The works in ~\cite{focken2002towards,mittal2002amulti} construct occupancy maps by using the foreground map after a background subtraction step.
Ground plane homographs are another technique introduced in \cite{khan2006multiview}, that generates a voting map from the foreground pixels in each view for occupancy map construction. Toward the probabilistic approach ~\cite{fleuret2007multicamera}, GMLP~\cite{9242263} jointly uses CNNs and Conditional Random Fields to model explicitly an occupancy volume map given detections estimated from multiple cameras. More recently~\cite{you2020real} (DMCT) propose deep learning to directly compute the occupancy volume by fusing feature maps extracted from CNNs at multi-camera views.




\myparagraph{Differences w.r.t.\ Previous Work}
Our work, denoted as LMGP, is at the intersection of single-view-based and centralized representation methods. We use single-camera tracklets but improve them by eliminating ID-switch errors using centralized representation concepts with multi-camera information derived from a novel 3D geometry-based occupancy map.
This factor sets us apart from competing approaches (Table \ref{table:reference-compare} Appendix). Moreover, we are the first to formulate a global lifted multicut method for multi-camera settings.
So far, lifted multicut was only applied to the single-camera setting~\cite{tang2015subgraph,tang2016multi,tang2017multiple}. We argue that our model is an elegant abstraction capturing the full range of interactions in multi-camera MOT.


\begin{figure*}[!hbt]
\centering
\includegraphics[width=1.0\textwidth]{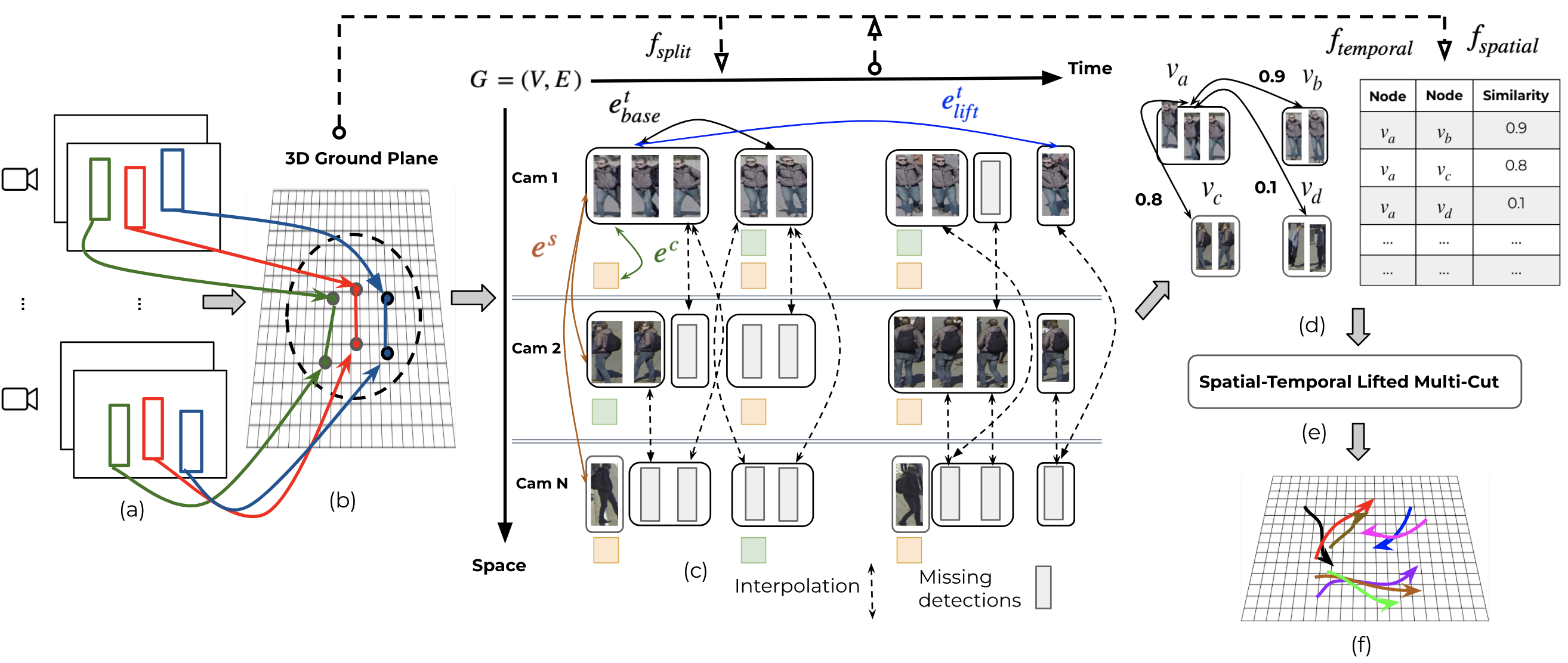}
\vspace{-0.3in}
\caption{
Illustration of our LMGP framework.
(a)~Input bounding boxes are given at each camera and each timepoint.
(b)~Bounding boxes observed from different cameras at the same time point are preliminary brought into correspondence through our 3D geometry based pre-clustering step.
(c)~A spatial-temporal tracking graph is constructed.
Nodes in the graph correspond to tracklets generated by a single-camera tracker that are split up at likely ID-switch positions based on features estimated from (b) step via an ID-Error predictor $f_{\mathrm{split}}$.
Edges correspond to possibly associations between tracklets with temporal edges $e^{t}$ (base and lifted for short and long range interaction in the same camera), spatial edges $e^{s}$ (different cameras, overlapping timeframes), and trajectory constraint edges $e^{c}$ (same camera, overlapping timeframes).
(d)~Association costs between nodes are computed between pre-processed tracklets using two networks $f_{\mathrm{temporal}}, f_{\mathrm{spatial}}$ for intra- and inter-camera edges in the tracking graph.
(e)~Tracklets in the tracking graph (c) are clustered together into trajectories via our lifted multicut optimization problem using affinity costs in (d).
(f)~3D coordinates of trajectories from (e) are generated.}
\label{fig:method-overview}
\vspace{-0.1in}
\end{figure*}

\section{Method}
Our tracking pipeline is illustrated in Figure~\ref{fig:method-overview}.
Below we describe each of its steps in detail, i.e.,\ pre-clustering for removing ID-switch errors and improving the subsequent affinity cost computation that is utilized in the global lifted multicut problem for computing multi-camera trajectories.

\myparagraph{Notation}
Before describing each part of our approach in detail, we introduce notation used throughout the paper. 
Let $B$ be the set of detections and $B^{t,j}$ the detections at timestep $t$ observed by camera $j$. 
Each detection $b \in B$ is observed by camera $\cam(b)$ and in timeframe $\dettime(b)$.
%
%
%
Each single camera tracklet $\tau$ consists of a set of bounding boxes at specific timepoints, i.e.\ $\tau = (b^\tau_1, b^{\tau}_2,..., b^{\tau}_{|\tau|})$, where $b^\tau_l$ is the $l$-th detection of trajectory $\tau$.
Tracklets only contain detections from a single camera, i.e.\ $\cam(b^\tau_1) =...= \cam(b^\tau_{|\tau|})$.
We extend the functions $\cam$ and $\dettime$ to tracklets by $\cam(\tau) = \cam(b_1)$ and $\dettime(\tau) = \{\dettime(b_1),\dettime(b_2),\ldots,\dettime(b_{|\tau|})\}$.
Detections of two tracklets that cover the same timepoint are denoted as
\begin{equation}
O(\tau, \tau') = \{(b,b') \in \tau \times \tau' :  \dettime(b) = \dettime(b')\} \, .
\end{equation}
%
We denote by $f$ a feature extractor that, given a bounding box $b$, produces an embedding vector $f(b)$ representing its appearance features.
$h$ denotes a map that takes a bounding box and computes the 3D coordinates of the foot point (center of bottom edge) on the ground plane $(z=0)$~\cite{hartley_zisserman_2004} (Section \ref{sec:transformations} Appendix).

\subsection{3D Geometry Based Pre-Clustering}
\label{sec:pre-clustering}
The pre-clustering step aims at bringing into correspondence detections of the same object observed by different cameras at each timeframe (Figure \ref{fig:method-overview}-b).
This enables us to overcome occlusions observed by a single camera.
In particular, if some object is occluded, we will be able to continue tracking the same object at different views (Figure~\ref{fig:teaser_crop}). Unlike prior works \citep{focken2002towards,mittal2002amulti,you2020real} that applied foreground subtraction or fusing image features from multiple cameras, our algorithm exploits 3D geometry constraints of detection projections of the same object. In particular, we project the bottom edge center of each bounding box to obtain its 3D coordinates (ground point) via map $h$.
Two detections observed by different cameras potentially belong to the same person if, after transformation to 3D, the Euclidean distance of the two ground points is less than the diameter of a typical person, which is approximately the human width average (Figure \ref{fig:2Dclustering}-a Appendix).  


The pre-clustering works as follows:
for each detection $b$, we consider the set of nearby detections $B^{t,j}(b) = \{b' \in B^{t,j}: \mathrm{dist}(h(b), h(b')) \leq r\}$ observed by the same camera $j = \cam(b)$ at the same timepoint $t = \dettime(b)$ with $\mathrm{dist}(,)$ and $r$ being Euclidean distance and radius to scan respectively. Likewise for camera $j' \neq \cam(b)$, we consider the set of detections $B^{t,j'}(b)$ observed by camera $j'$ close to the 3D-position $h(b)$.
We next compute a matching between detections of $B^{t,j}(b)$ and $B^{t,j'}(b)$ via a linear assignment problem~\cite{burkard1999linear} with costs being the Euclidean distance.
If $b$ is matched to 
a detection $b' \in B^{t,j'}(b)$ through the matching between $B^{t,j}(b)$ and $B^{t,j'}(b)$, and vice versa, if $b'$ is matched to $b \in B^{t,j}(b')$ through another matching between $B^{t,j'}(b')$ and $B^{t,j}(b')$, we record that match as it represents a confident connection. We denote the  resulting cluster for each detection $b$ as $C_b$.
The whole algorithm is detailed in Algorithm \ref{alg:pre-clustering} in the Appendix.

\myparagraph {Visible Detection Clustering}
Let $b \in B$ be a detection and $C_b$ be its cluster obtained after pre-clustering.
Since our algorithm uses only geometric coordinates of bounding boxes, the cluster $C_b$ can include both visible and occluded detections (Figure~\ref{fig:split-tracklet}-b).
Let $h_j$ be the 3D camera position of camera $j$ (Equation~\eqref{eq:camera_3D} in Section \ref{sec:transformations} Appendix).
First, given the detection $b$, we compute a detection $\visible(b)$ nearest to the camera by
\begin{equation}
\label{eq:occlusion-filtering}
\visible(b) = \begin{cases}\argmin_{b'} \dist(h(b'),h_j) \\
\text{s.t. } b' \in B^{\dettime(b),\cam(b)} : \iou(b',b) \geq 0.6\\
\end{cases}
\end{equation}
and then use it to refine the pre-cluster to contain only visible detections by
\begin{equation}
\label{eq:visible-detections-cluster}
C'_b = \{b' \in C_b : b' = \visible(b')\}\,.
\end{equation}

\subsection{Spatial-Temporal Tracking Graph}
\label{sec:spatial-temporal-tracking-graph}
We formulate a global spatial-temporal tracking graph $G = (V, E)$, where each node $v \in V$ corresponds to a tracklet $\tau$ in a single camera and edges represent data associations between tracklets across space and time (Figure \ref{fig:method-overview}-c).
A trajectory output will correspond to a cluster of nodes in the tracking graph $G$.
To benefit from current advances in single-camera MOT, each node (tracklet) at each camera is derived from a state-of-the-art tracker.
We use CenterTrack~\cite{zhou2020tracking} in our experiments, but it can be replaced by other trackers.

\begin{figure}[!hbt]
\centering
\includegraphics[width=0.5\textwidth]{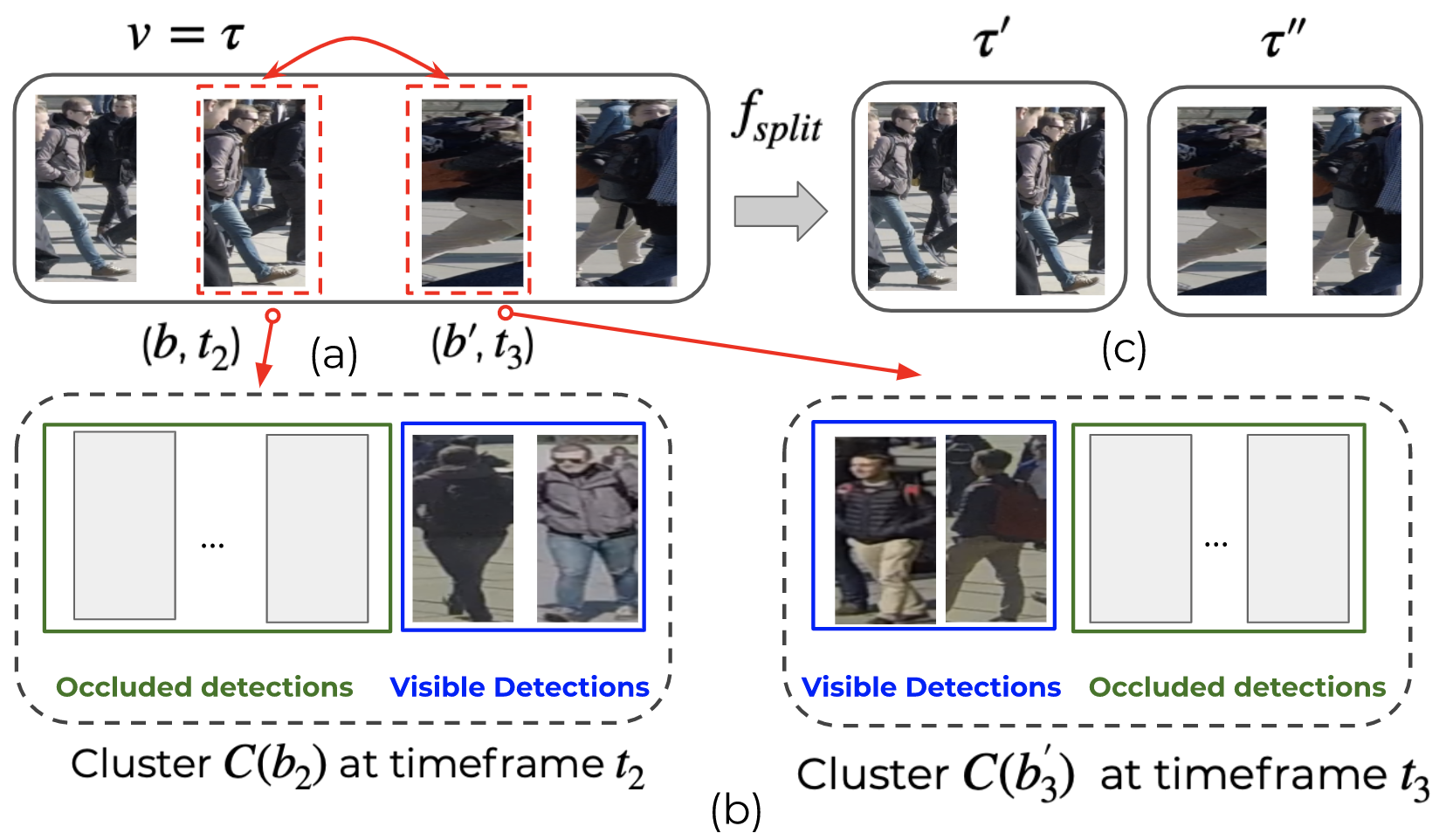}
\vspace{-0.3in}
\caption{(a) A tracklet (node) $\tau$ with ID-Switch error (two red dashed rectangles) in an initial tracking graph, (b) Using visible clusters of  detections at two consecutive frames, we can cut at the error positions, (c) Two new sub-tracklets $\tau',\, \tau''$  of distinct objects are generated.}
\label{fig:split-tracklet}
\end{figure}

While recent works \cite{xu2017cross,lan2020semi,he2020multi} directly compute affinity costs and solve the data association on the graph with nodes generated by single-camera trackers, we further correct ID-Switch errors in tracklet proposals (Figure~\ref{fig:split-tracklet}-a).
The ID switches in the original tracklets severely harm the total performance, especially in crowded or cluttered scenes.
To this end, we leverage bounding box correspondences from the pre-clustering and conduct the following steps.

\myparagraph{Feature Extension on Detection Clusters}
Given a pair of detections $b, b' \in B$ at different timepoints of potentially the same object and an embedding feature $f$ (we use DG-Net~\cite{zheng2019joint}), we aim to obtain robust association features by considering relations between detections in the visible cluster $C_{b}^{'}$ (Equation \ref{eq:visible-detections-cluster}). To this end, we first compute for all pairs of detections in $(\bar b, \bar b') \in C'_b \times C'_{b'}$ their cosine similarity w.r.t.\ features extracted by $f$, that is
$D_{\bar b, \bar b'} = \langle f(\bar b), f(\bar b') \rangle$.
Next, we solve a linear assignment problem between $C'_b$ and $C'_{b'}$ with costs $D_{\bar b, \bar b'}$ for each pair $(\bar b, \bar b') \in C'_b \times C'_{b'}$.
Let the set of matches be $M = \{(\bar b, \bar b') \in C'_b \times C'_{b'}: \bar b \text{ and } \bar b' \text{ are matched}\}$.
On the matches we estimate the following statistical similarities:
\begin{multline}
   c_{b,b'}^{\mathrm{best}} = \min_{(\bar b, \bar b') \in C'_b \times C'_{b'}} D_{\bar b, \bar b'}, \quad
   c_{b,b'}^{\mathrm{min}} = \min_{(\bar b, \bar b') \in M} D_{\bar b, \bar b'}, \\
   c_{b,b'}^{\mathrm{max}} = \max_{(\bar b, \bar b') \in M} D_{\bar b, \bar b'}, \quad
   c_{b,b'}^{\mathrm{mean}} = \frac{\sum_{(\bar b, \bar b') \in M} D_{\bar b, \bar b'}}{\abs{M}}, \\
   c_{b,b'}^{\mathrm{var}} = \sum_{(\bar b, \bar b') \in M} (D_{\bar b, \bar b'} - c_{b,b'}^{\mathrm{mean}})^2 \,.
   \label{eq:detection-cluster-features}
\end{multline}
\myparagraph{Splitting Tracklets}
\label{subsec:split-tracklets}
We now construct a network $f_{\mathrm{split}}$ (see Appendix \ref{appendix:network_architecture} for the architecture details) for correcting ID-Switch errors. Specifically, for each tracklet $\tau$,  $f_{\mathrm{split}}$
scans over all consecutive detections $(b,b') \subset \tau$ (Figure \ref{fig:split-tracklet}-a), takes their respective similarity values from Equations \ref{eq:detection-cluster-features} using the visible detections $C_{b}^{'},\, C_{b'}^{'}$ (Figure \ref{fig:split-tracklet}-b) and returns a probability score indicating whether or not they belong to the same tracklet.
$\tau$ is split into sub-tracklets at the predicted ID-switch error positions that become new nodes in our spatial-temporal tracking graph (Figure \ref{fig:split-tracklet}-c). 


\subsection{Learning Affinities with Multi-Camera Setting}
\label{sec:affinity-costs}
Given the tracking graph $G = (V, E)$ after pre-processing for ID-Switch error removal, we compute for pairs of tracklets $\tau=(b_1,\ldots,b_{\abs{\tau}})$ and $\tau' = (b'_1,\ldots,b'_{\abs{\tau'}})$ affinities representing the probability that both $\tau$ and $\tau'$ track the same object.
To this end, we consider standard cues based on motion information as in~\cite{9242263,xu2017cross,wen2017multi}.
We additionally propose novel appearance and 3D position-based similarity costs which harness image correspondences from our 3D pre-clustering step.
\vspace{-0.1in}
\subsubsection{Temporal Affinites }
\label{sec:temporal-edge-affinities}
Let $\tau$ and $\tau'$ be two tracklets, observed by the same camera and with $\dettime(b_{\abs{\tau}}) < \dettime(b'_1)$, i.e.\ following each other temporally.
We compute their similarity based on  motion and appearance.

\myparagraph{Forward/Backward Motion Affinities}
Motion extrapolation of $\tau$ to subsequent or extrapolation of $\tau'$ to previous timeframes can provide evidence of whether the two tracklets belong to the same object.
We adopt features from~\cite{xu2017cross,wen2017multi} for our settings by using the first/last $m$ frames from each tracklet to estimate an average velocity and then predict the forward velocity $\vec{v}_{\tau}$ of $\tau$ and the backward velocity $\overleftarrow{v}_{\tau'}$ of $\tau'$. The forward $c^{\mathrm{fw,t}}$ and backward $c^{\mathrm{bw,t}}$ affinity are computed by:
\begin{equation}
\label{eq:temporal-forward-backward-affinity-cost}
\begin{split}
 c^{\mathrm{fw,t}}(\tau, \tau') & =  \norm{h(b_{\abs{\tau}}) + \text{disp}(\tau,\tau') - h(b'_1)} \,.\\
c^{\mathrm{bw,t}}(\tau, \tau') & =  \norm{h(b'_{1}) - \text{disp}(\tau,\tau') - h(b_{\abs{\tau}})}  \,.
\end{split}
\end{equation}
where $\text{disp}(\tau,\tau') = \vec{v}_{\tau} \cdot t(\tau, \tau');\, t(\tau, \tau') = (\dettime(b'_1) - \dettime(b_{\abs{\tau}}))$. 


\myparagraph{Multi-view Appearance Affinities}
Tracklets belonging to the same object should share a similar appearance across time and cameras.
We measure this by computing
\begin{equation}
\label{eq:appearance-affinity}
        c^{\mathrm{app}}_{\mathrm{index}} (\tau, \tau') = \sum_{b \in \tau}\sum_{b' \in \tau'}  \frac{c_{\mathrm{index}}^{b,b'}}{\abs{\tau}\abs{\tau'}} \, , 
\end{equation}
where $\mathrm{index} \in \{\mathrm{best,min,max,mean,std}\}$ and each score $c_{\mathrm{index}}(b,b')$ is computed as in Equation \ref{eq:detection-cluster-features}. Note that we associate each detection with its visible image correspondences derived from the pre-clustering step in Equations \ref{eq:occlusion-filtering}, which sets us apart from prior works \cite{9242263,wen2017multi,lan2020semi} as both intra- and inter-camera information can be associated simultaneously.
An ablation study showing the benefit of this new cost can be found in Table \ref{tab:temporal-costs-performance} Appendix.
\vspace{-0.1in}
\subsubsection{Spatial Affinities}
\label{sec:spatial-edge-affinities}
For each pair of tracklets $\tau$ and $\tau'$ observed by different cameras and overlapping in time, we compute affinities based on similarity of motion and average 3D distance of their projected detections as in~\cite{xu2017cross,wen2017multi}.
We additionally propose  novel pre-clustering agreement based-similarity scores for each timestep on which they overlap.



\myparagraph{Forward/Backward Motion Affinities}
Similar to temporal forward resp.\ backward affinities, we will also exploit motion information in the spatial setting.
If, for example $\max\{\dettime(\tau)\} < \max\{\dettime(\tau')\}$,
we let $\vec{v}_{\tau}$ be the forward velocity of $\tau$.
The spatial forward affinity is defined as
\begin{equation}
    \label{eq:spatial-forward-affinity}
    c^{\mathrm{fw,s}}(\tau, \tau') = \norm{h(b_{\abs{\tau}}) + \vec{v}_{\tau} - h(b'_{\abs{\tau}+1})} \,.
\end{equation}
The spatial backward affinity $c^{bw,t}$ is defined analoguously.


\myparagraph{Average 3D Distance Affinity}
Given two tracklets of the same object, the bounding boxes at timepoints covered by both should have a small distance w.r.t.\ 3D projections.
We capture this property by defining the average 3D distance affinity as
\begin{equation}
\label{eq:average-3d-distance-affinity}
     c^{\mathrm{avg 3D}}(\tau, \tau') = \sum_{(b,b') \in O (\tau, \tau')}
    \frac{\norm{h(b) - h(b')}}{
   \abs{O(\tau,\tau')} 
    }\,.
\end{equation}

\myparagraph{Pre-clustering Agreement Affinity}
The pre-clustering of two tracklets of the same object on shared timesteps should coincide. We define a novel metric to quantify this by

\begin{equation}
\label{eq:pre-clustering-affinity}
c^{\mathrm{pc}}(\tau,\tau') = 
\sum_{(b,b') \in O(\tau,\tau')} \frac{p \cdot \mathbbmss{1}_{[C_b = C_{b'}]} + (1-p)\mathbbmss{1}_{[C_b \neq C_{b'}]} }{\abs{O(\tau,\tau')}} 
\end{equation}
for some prior probability $p$ (we choose $0.8$). 
The contributions of Equation \eqref{eq:pre-clustering-affinity} in capturing similarity among spatial tracklets can be seen in Table~\ref{tab:spatial-costs-performance} Appendix.

\subsection{Lifted Multicut for Multi-Camera Tracking}
\label{sec:lifted-multicut-optimization}
We first recapitulate the lifted multicut problem and use it to formulate the multi-camera MOT problem.
Our formulation extends the (lifted-) multicut works for single-camera MOT~\cite{tang2015subgraph,tang2016multi,tang2017multiple} to multiple cameras.

\myparagraph{Lifted Multicut} 
The multicut optimization problem~\cite{chopra1993partition} is to partition nodes of a given graph $G=(V,E)$ with edge weights $c: E \rightarrow \R$ into a number of clusters $\Pi = (\Pi_1,\ldots,\Pi_k)$, where the number of clusters $k$ is determined as part of the optimization problem.
The clusters form a partition of $V$, i.e.\ $\Pi_i \cap \Pi_j = \varnothing$ $\for i \neq j$ and $\Pi_1 \cup \ldots \cup \Pi_k = V$ where each component $\Pi_i  \in \Pi$ implies a connected subgraph of $G$.
A multicut is an edge indicator vector $y: {E} \rightarrow \{0,1\}$ defined  by $y_e = 0$ iff the endpoints of $e$ are in the same component. 
The multicut problem can be written as an integer linear program:
\begin{equation}
        \min_{y \in \{0,\,1\}^{E}} \sum_{e \in E}c_{e}\,y_{e}
\end{equation}
\vspace{-0.1in}
\begin{equation}
\label{eq:multi-cut-constraint}
        \mathrm{s.t.\ } \forall C \in \mathrm{cycles}(G),\ \forall e \in C: y_{e} \leq \sum_{\bar{e} \in C  \setminus \{e\}}\, y_{\bar{e}}.
\end{equation}

The lifted multicut problem~\cite{keuper2015lifted} augments the original multicut problem by introducting a second set of lifted edges $E'$ together with lifted edge costs $c' : E' \rightarrow \R$.
For any edge $e' = (i,j) \in E'$ and clusters $(\Pi_1,\ldots,\Pi_k)$, a lifted edge indicator vector is defined as
\begin{equation}
\label{eq:lifted_constraints}
    \hspace{-0.15in}y_{e'} = 0 \Leftrightarrow \exists l \text{ and a path }P 
    \in ij\text{-paths}(G)
    \text{ s.t. } P \in \Pi_l.\hspace{-0.1in}
\end{equation}
In words, the label of lifted edge is $0$ iff there exists a path connecting its endpoints through base edges in $E$ and through nodes that all are in a single cluster.

\myparagraph{Spatial-Temporal Lifted Tracking Graph} We now propose our lifted multicut for multiple object tracking given a spatial-temporal graph. Specifically, three types of edges are employed: temporal, spatial, and constraint edges (Figure \ref{fig:method-overview}-c). The temporal edges $E^{t}_{t_{\mathrm{max}}}$ connect tracklet nodes observed by the same camera at different timesteps up to some maximal time threshold $t_{\mathrm{max}}$ (10 seconds in our experiments), spatial edges $E^{s}$ connect tracklets observed by different cameras at overlapping timeframes, and tracklet nodes that are observed by the same camera and have overlapping timeframes are forbidden to end up in the same trajectory through constraint edges $E^c$. The formulations for these edge sets are:
\begin{equation}
\resizebox{.5 \textwidth}{!}{$
    \begin{split}
        E^{t}_{t_{\mathrm{max}}} & = \left\{ (\tau,\tau') :
        \begin{array}{c}
        \cam(\tau) = \cam(\tau'), \\
         \max\{\dettime(\tau')\} - \min\{\dettime(\tau)\} \in (0, t_{max}]
        \end{array}
        \right\} \\
        E^{s} & = \left\{ (\tau,\tau') : 
        \begin{array}{c}\cam(\tau) \neq \cam(\tau'), \\
        \dettime(\tau) \cap \dettime(\tau') \neq \varnothing
        \end{array}
        \right\}\\
        E^c &= \left\{ (\tau, \tau') \in V \times V:
        \begin{array}{c}
        \cam(\tau) = \cam(\tau'), \\
        \dettime(\tau) \cap \dettime(\tau') \neq \varnothing 
        \end{array}
        \right\}\,. 
    \end{split}$}
\end{equation}
Given the above edges, we divide them into 
base edges $E$ and lifted edges $E'$:
\begin{equation}
    E = E^{t}_{\mathrm{5sec}}  \cup E^c \cup E^s;
    E' = E^{t}_{\mathrm{10sec}} \backslash E^t_{\mathrm{5sec}}\,.
\end{equation}
That is, all edges such that tracklets have temporal distance less than $5$ seconds are base edges ($e^{t}_{\mathrm{base}}$), while temporal edges with larger time distance $t_{\mathrm{max}}$ are lifted ones ($e^{t}_{\mathrm{lift}}$).
The lifted edges are used here to incorporate long-range interactions of objects, i.e.\ handling the case when an object disappears due to occlusion and reappears again.

The edge costs $c_{e^{t}}$ and $ c_{e^{s}}$ for temporal $e^{t} \in E^{t}_{\mathrm{t_{max}}}$ and spatial edges $e^{s} \in E^{s}$ in our framework are produced by two neural networks $f_{\mathrm{temporal}}$ and $f_{\mathrm{spatial}}$, which take input features described in Section \ref{sec:affinity-costs} and return a similarity score (Subsection \ref{sec:computing-affinity} Appendix). 
We also assign a large negative value $c_{e^{c}} = M \ll 0$ for each $e^{c} \in E^c$ to guarantee that clusters are trajectories.

\myparagraph{Lifted Multicut Formulation}
We state our lifted multicut optimization for the multi-camera tracking as: 
\begin{equation}
\begin{gathered}
    \min\limits_{y \in \{0,1\}^{E \cup E'}}  \sum\limits_{e^{t} \in E^{t}_{t_{\mathrm{max}}}}c_{e^{t}}\,y_{e^{t}} + \sum\limits_{e^{s} \in E^{s}}c_{e^{s}}\,y_{e^{s}} + \sum\limits_{e^{c} \in E^{c}}M\,y_{e^{c}}\\
    \text{s.t. }  y \text{ obeys Equations }\eqref{eq:multi-cut-constraint} \text{ and }\eqref{eq:lifted_constraints}.
    \end{gathered}
    \label{eq:proposed-lmp}
\end{equation}
Since this formulation is NP-Hard \cite{keuper2015lifted}, we resort to efficient heuristic solvers presented in Section \ref{sec:solver-lmp} Appendix.

In summary, our formulation has the following advantages. First, we optimize intra-camera (temporal edges) and inter-camera (spatial edges) simultaneously. Additionally, long-range interactions are also incorporated through lifted-edges in $E'$.
This results in a solution informed by cues from all cameras at once.
Second, the optimal number of trajectories is determined during the optimization. Third, our nodes in the spatial-temporal graph are tracklets, resulting in decreased execution time as the hypothesis space is significantly reduced.

\section{Experimental Analysis}
Below we detail our experimental setup and results, including implementation details of our approach, considered datasets and baselines, quality metrics, and ablations.


\subsection{Implemention Details}

\myparagraph{Tracking Graph}
We use CenterTrack~\cite{zhou2020tracking}, one of current state-of-the-art single-camera trackers, to create tracklets for each camera.
These tracklets will be aligned again with the provided public detections to eliminate duplicated ones and retain detections that were not tracked, e.g.\ due to occlusion.
We apply DG-Net~\cite{zheng2019joint} for extracting embedding vectors for detections and train it on visible detections (Eq.~\ref{eq:occlusion-filtering}) obtained by running the proposed pre-clustering algorithm over training video sequences.
Finally, we train three basic multi-layer deep networks for the three networks $f_{\mathrm{split}}$, $f_{\mathrm{spatial}}$, and $f_{\mathrm{temporal}}$ which handle splitting up initial tracklets, generating affinity costs for spatial and temporal edges respectively.

\myparagraph{Lifted Multicut Solver}
For solving the proposed lifted multicut problem (Eq. \ref{eq:proposed-lmp}), we use the efficient GAEC+KLj heuristic solver from~\cite{keuper2015lifted}.
We adapt a two-stage optimization approach. First, the initial tracklets are used as nodes in the tracking graph and compute trajectories through the lifted multicut.
Second, we initialize the computed trajectories again as nodes in the tracking graph and recompute costs to obtain the final trajectories. The second step improves the trajectories computed in the first pass by allowing to compute costs on longer trajectories and joining trajectories together where a connection was not initially detected. For a long-time video, this process can be iterated until converged. More details of our implementations are described in Subsection \ref{sec:implement} Appendix.

\subsection{Datasets \& Metrics}
We perform experiments on three datasets with a wide range of different camera configurations, densities, and video/bounding box qualities.
For all datasets, we use the provided detections for a fair comparison.

\begin{packed_description}
\item[- \normalfont\textit{WILDTRACK}~\cite{chavdarova2018wildtrack}:]
The largest-scale dataset for the multi-camera setting is currently with a dense group of $313$ pedestrians standing and walking.
There is a total of $400$ frames observed from seven cameras that are annotated with 3D positions.
The first $360$ of those frames is used for training and the rest for testing. 
\item[- \normalfont\textit{Campus}~\cite{xu2016multi}:] We have chosen the two sequences \textit{Garden~1} and \textit{Parking~Lot} which had camera calibration parameters and a ground plane for 3D-projection. There are 15 - 25 pedestrians in each video observed by four cameras captured at a 30fps rate.
For each video, we use the first $10\%$ for training and the remaining frames for testing. 
\item[- \normalfont\textit{PETS-09}~\cite{ellis2009pets2009}:]
Contains three sequences with increasing level of difficulty:
low density (S2.L1) - $19$ objects in $795$ frames, medium density (S2.L2) - $43$ pedestrians spreading in $436$ frames and high density (S2.L3) - $44$ pedestrians moving together in $240$ frames.
While PETS-09 is not as dense as WILDTRACK, its main challenge lies in its poor video acquisition conditions with cameras far away from targets and low-quality bounding boxes. 
\end{packed_description}
\vspace{-0.1in}
We report results for the following metrics:
\begin{packed_description}
\item[- \normalfont MOTA~\cite{bernardin2008evaluating}:] the multiple objects tracking accuracy measuring a number of false negatives (FN), false positives (FP), and identity switches (IDs) focusing on the coverage of detections. 
\item[- \normalfont MOTP~\cite{bernardin2008evaluating}:]
the multiple object tracking precision penalizes the overall dissimilarity between true positives and the ground truth objects.
\item[- \normalfont IDF1~\cite{ristani2014tracking}:] 
measure through the F1 score agreement between computed and ground truth trajectories.
%
%
%
\item[- \normalfont MT \cite{li2009learning}:] number of mostly tracked objects for at least $80\%$ of its life span.
\item[- \normalfont ML \cite{li2009learning}:] number of mostly lost objects for at most $20\%$ of its life span.
\end{packed_description}
\subsection{Algorithms}
We compare against state-of-the-art baselines for each respective dataset.
To study the effectiveness of our approach and judge the effectiveness of each proposed component, we run experiments on multiple configurations. All algorithms are listed below.
\begin{packed_description}
\item[\normalfont\texttt{Baselines}:]
For each dataset, we compare against the most recent approaches for which we found experimental results.
In particular, we compare against KPS-DO,  KSP-DO-ptrack  \citep{chavdarova2018wildtrack}, GLMB-YOLOv3, GLMB-DO  \citep{9242263}, DMCT, DMCT Stack \cite{you2020real}, HJMV~\cite{hofmann2013hypergraphs}, STVH~\cite{wen2017multi}, MLMRF~\cite{lan2020semi}, HCT~\cite{xu2016multi}, STP~\cite{xu2017cross}, TRACTA~\cite{he2020multi} and DyGLIP~\cite{quach2021dyglip}.
All baseline performances are taken from the respective original papers. 
\item[\normalfont Our configurations are:]
\item[\normalfont\texttt{- LMGP w/o Pre-Clustering}:] see Section \ref{sec:pre-clustering}.
\item[\normalfont\texttt{- LMGP w/o Tracklet Split}:] no tracklet splitting using information from pre-clustering (Section \ref{sec:spatial-temporal-tracking-graph}, Splitting Tracklets).
\item[\normalfont\texttt{- LMGP w/o Enhanced Affinities}:] do not use pre-clustering in affinity cost computation (Eq. \ref{eq:appearance-affinity},
\ref{eq:pre-clustering-affinity}).
\item[\normalfont\texttt{- LMGP}:] our approach with everything enabled.
For the \textit{Campus} dataset we also provide as comparison \texttt{LMGP-DeepSort} using DeepSort~\cite{wojke2017simple} for generating tracklets.
Note that CenterTrack outperforms DeepSort on single camera benchmarks.
\end{packed_description}

\subsection{Results}
\vspace{-0.1in}
\begin{table}[!hbt]
\scalebox{0.63}{
\begin{tabular}{c|c|c|c|c|c|c|c}
\Xhline{2\arrayrulewidth}
Method& IDF1 $\uparrow$ & MOTA  $\uparrow$ & MT  $\uparrow$   & ML  $\downarrow$  & FP $\downarrow$ & FN $\downarrow$ & IDs $\downarrow$ \\ \Xhline{2\arrayrulewidth}
\small KSP-DO  \citep{chavdarova2018wildtrack}        & 73.2          & 69.6                               & 28.7          & 25.1        & 1095        & 7503        & 85           \\ 
\small KSP-DO-ptrack \citep{chavdarova2018wildtrack} & 78.4          & 72.2                               & 42.1         & 14.6         & 2007        & 5830        & 103          \\ 
\small GLMB-YOLOv3      \citep{9242263}       & 74.3          & 69.7                               & 79.5          & 21.6         & 424         & 1333        & 104          \\ 
\small GLMB-DO     \citep{9242263}      & 72.5          & 70.1                               & 93.6         & 22.8         & 960         & 990         & 107          \\ 
\small DMCT  \citep{you2020real}                   & 77.8          & 72.8                               & 61.0            & 4.9         & 91          & 126         & 42           \\ 
\small DMCT Stack  \citep{you2020real}              & 81.9          & 74.6                               & 65.9          & 4.9          & 114         & 107         & 21           \\ 
\hdashline
\small LMGP w/o Pre-Clustering             & 79.4     & 73.6                        & 76.4      & 24.3     & 570   & 506  & 76    \\ 
\small LMGP w/o Tracklet Split & 92.4    & 89.7                        & 92.6      & 8.3     & 285  & 119  & 45   \\
\small LMGP w/o Enhanced Affinities & 95.1   & 94.8                         & 95.5       & 5.7      & 95  & 103  & 27   \\
\small LMGP            & \textbf{98.2} & \textbf{97.1}                      & \textbf{97.6} & \textbf{1.3} & \textbf{71} & \textbf{7}  & \textbf{12}  \\ \Xhline{2\arrayrulewidth}
\end{tabular}}
\vspace{-0.1in}
\caption{Our LMGP performance compared to state-of-the-art baselines on \textit{WILDTRACK}.}
\label{tab:multi-wildtrack}
\end{table}
\vspace{-0.2in}
\begin{table}[!hbt]
  \centering
  \scalebox{0.62}{
  \begin{tabular}{c|c|c|c|c|c|c}
\Xhline{2\arrayrulewidth}
   \multicolumn{1}{c|}{Sequence}   & \multicolumn{1}{c|}{Method} & \multicolumn{1}{c|}{MOTA $\uparrow$} & \multicolumn{1}{c|}{MOTP $\uparrow$} & \multicolumn{1}{c|}{MT $\uparrow$} & \multicolumn{1}{c|}{ML $\downarrow$} &
     \multicolumn{1}{c}{IDs $\downarrow$}\\ \Xhline{2\arrayrulewidth}
     \multirow{5}{*}{\STAB{\rotatebox[origin=c]{0}{S2-L1}}}
     & HJMV  \citep{hofmann2013hypergraphs}                   &   91.7                        &     79.4                      &  94.7                         &  \textbf{0.0 }                      & 45 \\
     & STVH   \citep{wen2017multi}              &   95.1                        &          79.8                  &  \textbf{100.0}                         &  \textbf{0.0 }                      & 13  \\
     & MLMRF      \citep{lan2020semi}               &  96.8                         &       79.9                    &  \textbf{100.0}                        &   \textbf{0.0 }                    & \textbf{2} \\ 
     \hdashline
     & LMGP w/o Pre-Clustering                     &                  97.5       &        79.6                     &  96.3                         &   \textbf{0.0}                       & 6 \\
     & LMGP w/o Tracklet Split                      &    97.3                       &         82.1                  &                       97.2    &    \textbf{0.0}                      & 6 \\
     & LMGP w/o Enhanced Affinities                     &  97.6                         &    82.1                       &     98.1                     &        \textbf{0.0}                & 4 \\
     & LMGP                      &   \textbf{ 97.8}                     &          \textbf{82.4}                  &      \textbf{100.0}                     &   \textbf{0.0 }                      & \textbf{2} \\\Xhline{2\arrayrulewidth}

          \multirow{5}{*}{\STAB{\rotatebox[origin=c]{0}{S2-L2}}}
     & HJMV  \citep{hofmann2013hypergraphs}                &   58.9                       &    66.0                         &  30.2                         &  2.3                      & 388  \\
     & STVH   \citep{wen2017multi}                       &     65.2                       &     61.8                    &  44.2                         & \textbf{ 0.0 }                       & 249  \\
     & MLMRF      \citep{lan2020semi}                  &      \textbf{ 72.1 }                  &     58.3                   &    \textbf{72.1 }                     &     2.3                     & 142 \\ 
     \hdashline
     & LMGP w/o Pre-Clustering                      &                  66.8        &            64.3                &   67.4                        &    2.1                      &  158 \\
     & LMGP w/o Tracklet Split &    70.1                       &    68.5                       &             67.2              &    2.1                      & 97 \\
     & LMGP w/o Enhanced Affinities                      &  71.6                       &  \textbf{73.2 }                        &   68.2                       &   2.0                       & 81 \\
     & LMGP                               &  70.4   &    \textbf{73.2}                          &       69.6                    &      1.7                  & \textbf{75}\\\Xhline{2\arrayrulewidth}
          \multirow{5}{*}{\STAB{\rotatebox[origin=c]{0}{S2-L3}}}
& HJMV  \citep{hofmann2013hypergraphs}              &   40.2                        &   49.5               &  29.6                         &  25.0                     & 123 \\
     &STVH   \citep{wen2017multi}                & 49.8                        &        63.0                   &    29.6                       &      20.5                    &   92\\
     & MLMRF      \citep{lan2020semi}                & \textbf{54.4}                         &    54.9                         &   36.4                        &      6.8                    & 82 \\
     \hdashline
     & LMGP w/o Pre-Clustering                     &     49.1                     &      55.7                      & 33.2                          &    13.6                      &  92 \\
     & LMGP w/o Tracklet Split                       & 52.7                          &   63.6                         &      33.4                     &   11.2                       & 86 \\
      & LMGP w/o Enhanced Affinities                      &         53.5                 &     64.7                     &    31.8                      &          8.7                & 82 \\
     & LMGP                      &              \textbf{54.4 }             &           \textbf{66.5}               &   \textbf{40.2}                        &  \textbf{5.3 }                       & \textbf{78}\\ 
\Xhline{2\arrayrulewidth}
\end{tabular}}
\vspace{-0.1in}
\caption{Our LMGP performance compared to other baselines in \textit{PETS-09}.}
\label{tab:pets09}
\end{table} 
\vspace{-0.2in}
\begin{table}[!hbt]
  \centering
  \scalebox{0.65}{
  \begin{tabular}{c|c|c|c|c|c}
\Xhline{2\arrayrulewidth}
   \multicolumn{1}{c|}{Sequence}   & \multicolumn{1}{c|}{Method} & \multicolumn{1}{c|}{MOTA $\uparrow$} & \multicolumn{1}{c|}{MOTP $\uparrow$} & \multicolumn{1}{c|}{MT $\uparrow$} & \multicolumn{1}{c}{ML $\downarrow$} \\ \Xhline{2\arrayrulewidth}
     \multirow{5}{*}{\STAB{\rotatebox[origin=c]{0}{Garden 1}}}
     & HCT  \citep{xu2016multi}                   &   49.0                       &     71.9                     &  31.3                         &  6.3                     \\
     & STP   \citep{xu2017cross}              &   57                       &          75                  &  -                       & -                \\
     & TRACTA      \citep{he2020multi}               &  58.5                        &       74.3                    &  30.6                        &   1.6              \\ 
      & DyGLIP      \citep{quach2021dyglip}               &  71.2                      &       91.6                   &  31.3                        &   \textbf{0.0 }                   \\ 
     \hdashline
      & LMGP-DeepSort                    &   75.6                     &          93.4                 &      46.7                     &   1.6                      \\
     & LMGP                      &   \textbf{ 76.9}                     &          \textbf{95.9}                  &      \textbf{62.9}                     &   1.6                      \\\Xhline{2\arrayrulewidth}
          \multirow{5}{*}{\STAB{\rotatebox[origin=c]{0}{Parking Lot}}}
     & HCT  \citep{xu2016multi}                &   24.1                      &    66.2                         &  6.7                         &  26.6                     \\
     & STP   \citep{xu2017cross}                       &     28                       &     68                    & -                      & -                \\
     & TRACTA      \citep{he2020multi}                  &     39.4              &     74.9                   &   15.5                   &     10.3                 \\ 
       & DyGLIP      \citep{quach2021dyglip}               &  72.8                         &       \textbf{98.6}                  &  26.7                        &   \textbf{0.0 }                   \\ 
     \hdashline
      & LMGP-DeepSort                             &  76.7   &    98.0                       &       51.7                    &    5.1         \\
     & LMGP                               &  \textbf{78.1}   &    97.3                       &       \textbf{62.1}                    &     \textbf{0.0  }          \\\Xhline{2\arrayrulewidth}
  \end{tabular}}
\vspace{-0.1in}
\caption{Our LMGP performance compared to other baselines on \textit{Campus}.}
\label{tab:campus}
\end{table} 
\vspace{-0.2in}
We report quantitative results for \textit{WILDTRACK} in Table~\ref{tab:multi-wildtrack}, for \textit{PETS-09} in Table~\ref{tab:pets09}  and  for \textit{Campus} in Table~\ref{tab:campus}.
Some qualitative results are presented in Figures~\ref{fig:qualitative-results},\,\ref{fig:qualitative-results-campus} Appendix.

On \textit{WILDTRACK} we obtain almost perfect metric scores with \texttt{LMGP}. Our results on \textit{Campus} significantly outperform the state-of-the-art on both sequences even when using weak tracklets (DeepSort).
On \textit{PETS-09} we achieve comparable results.
We argue that the differing performance of our algorithms is mostly due to poor bounding boxes and camera calibration for \textit{PETS-09} and, to a lesser extent, for \textit{Campus}.
\subsection{Efficacy of Individual Components \& Ablations}
\label{sub:ablation-study}
To assess the performance of individual parts in our approach and their contribution to our overall performance, we give more experimental details.

\myparagraph{Ablations w.r.t.\ Pre-Clustering/Enhanced Affinities/Tracklet Splitting}
In Tables~\ref{tab:multi-wildtrack} and~\ref{tab:pets09} we report results of the ablated versions \texttt{LMGP w/o Pre-Clustering}, \texttt{LMGP w/o Tracklet Split} and \texttt{LMGP w/o Enhanced Affinities} of our solver. 
In most cases (except for \textit{S2-L1} of \textit{PETS-09}, where nearly perfect results can be obtained with any baseline), we see significant improvements w.r.t.\ all ablated versions of our solvers, validating the efficacy of all steps. 
\textit{The greatest performance drop can be observed by turning of the pre-clustering}, showing its important role for obtaining improved results.
More fine-grained ablations for the enhanced affinity costs can be found in Table~\ref{tab:temporal-costs-performance} and~\ref{tab:spatial-costs-performance} Appendix.

\begin{figure}[!hbt]
\centering
\includegraphics[width=0.38\textwidth]{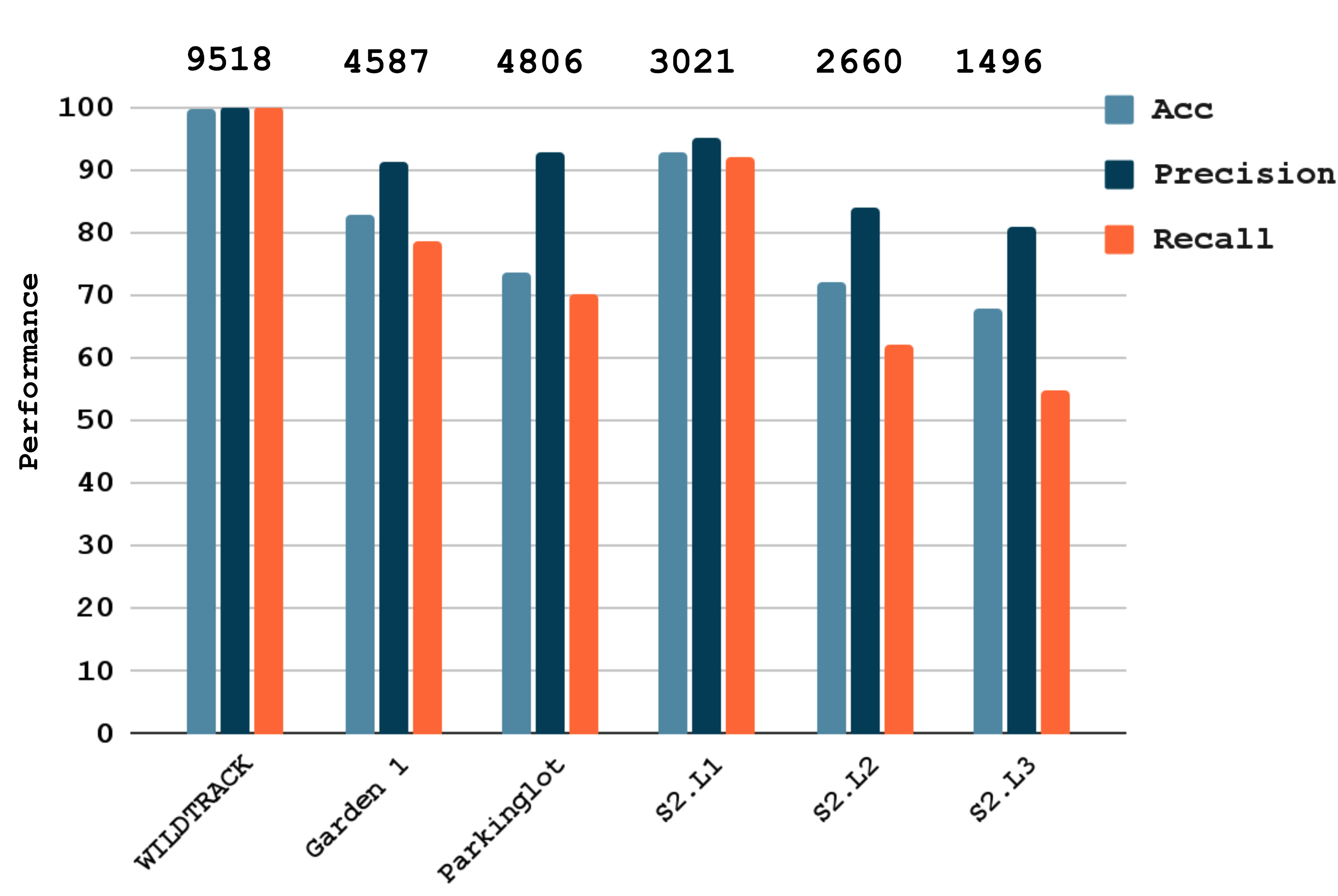}
\vspace{-0.15in}
\caption{Our pre-clustering performance measured by accuracy, precision and recall w.r.t.\ correctly estimated correspondences between detections across cameras. On the top we give the total detections of objects in each dataset.
}
\label{fig:result-2Dclustering}
\vspace{-0.2in}
\end{figure}

\myparagraph{Pre-Clustering Accuracy}
In Figure~\ref{fig:result-2Dclustering}, we report performance of our Pre-Clustering (Section~\ref{sec:pre-clustering}) for all considered datasets.
Since our Pre-Clustering is purely based on geometry, we also see experimentally that its performance is dependent upon the accuracy of bounding box coordinates.
In the high-quality \textit{WILDTRACK} dataset, we obtain almost perfect results.
For \textit{Campus} and \textit{PETS-09} we achieve precision higher than $80\%$.
We argue that the \textit{confident connection} in our pre-clustering helps in these more challenging and noisy settings.
More details on the reduction of ID-switch errors can be found in Table~\ref{tab:center-tracktor} Appendix.
\vspace{-0.15in}

\myparagraph{Joint Optimization Model}
In order to assess the performance of our joint spatio-temporal optimization model (Eq.\ref{eq:proposed-lmp}) as compared to a stage-wise optimization, we provide an experiment in  Table~\ref{tab:joint-spatial-temporal-optimize}.
The variant \texttt{LMGP w/o spatial edges} first solves an ordinary single view MOT problem. After obtaining computed single-camera trajectories, they are linked across cameras in a second step. To validate the effect of lifted long-range edges vs. a simpler model without them, we also provide an ablation \texttt{LMGP w/o lifted edges}. These ablations are tested on \textit{WILDTRACK} and \textit{S2-L1} of \textit{PETS-09}. The results 
indicate that significant improvement can be gained by optimizing jointly over temporal and camera affinities, especially in dense scenes like WILDTRACK. Also, the long-range edges contribute to better performance in both cases.
\begin{table}[!hbt]
\vspace{-0.1in}
  \centering
  \scalebox{0.6}{
  \begin{tabular}{c|c|c|c|c|c|c}
\Xhline{2\arrayrulewidth}
    \multicolumn{1}{c|}{\small Dataset} & \multicolumn{1}{c|}{\small Method } & \multicolumn{1}{c|}{\small MOTA $\uparrow$} & \multicolumn{1}{c|}{\small IDF1 $\uparrow$} & \multicolumn{1}{c|}{\small MT $\uparrow$} &
    \multicolumn{1}{c|}{\small ML $\downarrow$} &
    \multicolumn{1}{c}{\small IDs $\downarrow$}\\ \Xhline{2\arrayrulewidth}
     \multirow{2}{*}{\STAB{\rotatebox[origin=c]{0}{\small{\textit{WILDTRACK}} 
     }}}
     & \small {LMGP full}                      &   \textbf{97.1}                        &  \textbf{98.2}                         &  \textbf{97.6}                         &  \textbf{1.3}                   &  \textbf{12} \\
       & \small{LMGP w/o lifted edges}                      &   95.4                        &  96.4                         &  93.6                         &  2.7                  & 41  \\
     & \small{LMGP w/o spatial edges}                      &   93.2                        &  94.1                         &  91.7                         &  6.9                  & 85  \\
 \hline
          \multirow{2}{*}{\STAB{\rotatebox[origin=c]{0}{\small{\textit{S2-L1}}}}}
     & \small{LMGP full}                      &     \textbf{97.8}                      &  \textbf{82.4}                         &  \textbf{100.0}                        &  \textbf{0.0}                   &   \textbf{2}  \\
       & \small{LMGP w/o lifted edges}                      &  96.2                        &  81.1                         &  98.7                         &  1.3                    & 5 \\
     & \small{LMGP w/o spatial edges}                      &  95.6                        &  80.2                         &  98.7                         &  1.3                    & 8 \\
\Xhline{2\arrayrulewidth}
  \end{tabular}}
  \vspace{-0.1in}
  \caption{An ablation study of different edge types}
  \label{tab:joint-spatial-temporal-optimize}
\end{table} 
\vspace{-0.25in}
\subsection{Discussion}
We have demonstrated that nearly perfect multiple object tracking results can be obtained in crowded scenes given the right conditions (as for \textit{WILDTRACK}), i.e.\ high bounding box quality and a large enough number of well-calibrated cameras observing the same scene.
Even when these conditions are not met, as in \textit{Campus} and \textit{PETS-09} which use old detectors, our \texttt{LMGP} still delivers better results than competitors in most cases.
Specifically, the \textit{pre-clustering is crucial in our framework} because it allows exploiting multi-camera information through repairing efficiently trajectories computed by single-camera trackers and can be used to enhance affinity costs.
Also, our lifted multicut model jointly optimizing over inter- and intra-camera affinities, short- and long-range interactions is essential and can effectively correct erroneous associations and continue trajectories which would be lost when only a single camera is used and severe occlusions are present.
\vspace{-0.1in}
\section{Conclusion}
%
We have shown that integrating single view-based approaches and centralized representation-based methods for multi-camera tracking can lead to improvements w.r.t.\ strategies that fall into only one of these paradigms. Given a good enough input data, this strategy can deliver almost optimal results even in crowded scenes.
We conjecture that in noisier settings, significantly better results can be achieved by making our 3D pre-clustering more robust, e.g., by employing it in an end-to-end training framework \citep{xu2020train,braso2020learning}.



\vspace{-0.1in}
\section{Acknowledgements}
This research is sponsored by the XAINES and pAItient projects (BMBF grant no. 01IW20005 and BMG grant no. 2520DAT0P2 respectively).
\setlength{\bibsep}{0pt}

{\small
\bibliographystyle{ieee_fullname}
\bibliography{literature}
}
\newpage
\appendix
\section*{\Large Supplementary Material}
The appendix complements our method by a detailed comparison with prior works, transformations between image coordinates and the world coordinates, a pseudocode representation of the proposed 3D geometry pre-clustering step, an additional sampling strategy for spatial-temporal edge costs, our lifted multicut solver, and a method to create tracking predictions in 3D space. Along with it, we present the performance of pre-processing the spatial-temporal tracking graph and ablation studies on the final edge costs accuracy, which depends on our pre-clustering algorithm. Lastly, we show the performance of LMGP compared with state-of-the-art single-camera methods, computational time, and some qualitative tracking results on WILDTRACK and Campus dataset. \\

\section{LMGP Compared with Other Multi-Camera Trackers}
We present in Table \ref{table:reference-compare} the main differences between our LMGP method with other baselines. In summary, we utilize image correspondences for each object at each timeframe as the \textit{Centralized Representation} strategy to correct ID-Switch errors generated by tracklets in \textit{Single View-based methods}. On the contrary, using nodes as tracklets allows us to reduce the computational cost in the association step significantly, implying our tracker runs reasonably fast and, therefore, be feasible to deploy it in real-world tracking applications (Table \ref{tab:running-time}).

\begin{table}[!hbt]
    \centering
    \scalebox{0.9}{
    \begin{tabular}{cccccc}
\Xhline{2\arrayrulewidth}
Method & \multicolumn{1}{c}{\begin{tabular}[c]{@{}c@{}}Single-Camera\\ Tracklets\end{tabular}} & \multicolumn{1}{c}{\begin{tabular}[c]{@{}c@{}} Centralized \\ Representation\end{tabular}} & Online \\ \Xhline{2\arrayrulewidth} 
    \small {LMGP (Ours)} & \checkmark & \checkmark & \\
    \small{MLMRF} \cite{lan2020semi} &\checkmark & &\checkmark \\ 
    \small{STVH} \cite{wen2017multi} & \checkmark &\\
    \small{DMCT} \cite{you2020real} & & \checkmark & \checkmark \\
    \small{GLMB} \cite{9242263} &  & \checkmark & \checkmark \\      
     \small{DyGLIP} \cite{quach2021dyglip} & \checkmark &  & \checkmark \\ \Xhline{2\arrayrulewidth}
    \end{tabular}}
    \vspace{0.05in}
    \caption{
    Comparison of our LMGP w.r.t.\ recent multi-camera trackers.
    Single camera tracklets signifies for trackers whose inputs can be used as tracklets. Centralized representation refers to usage of relationships among detections at each timeframe. Online indicates for only current frames are used. 
    }
    \label{table:reference-compare}
\end{table}

\section{Transformation From Camera Coordinates to  Word Coordinates on The Ground Plane}
\label{sec:transformations}

\begin{figure}[!htb]
\centering
    \includegraphics[width=0.5\textwidth]{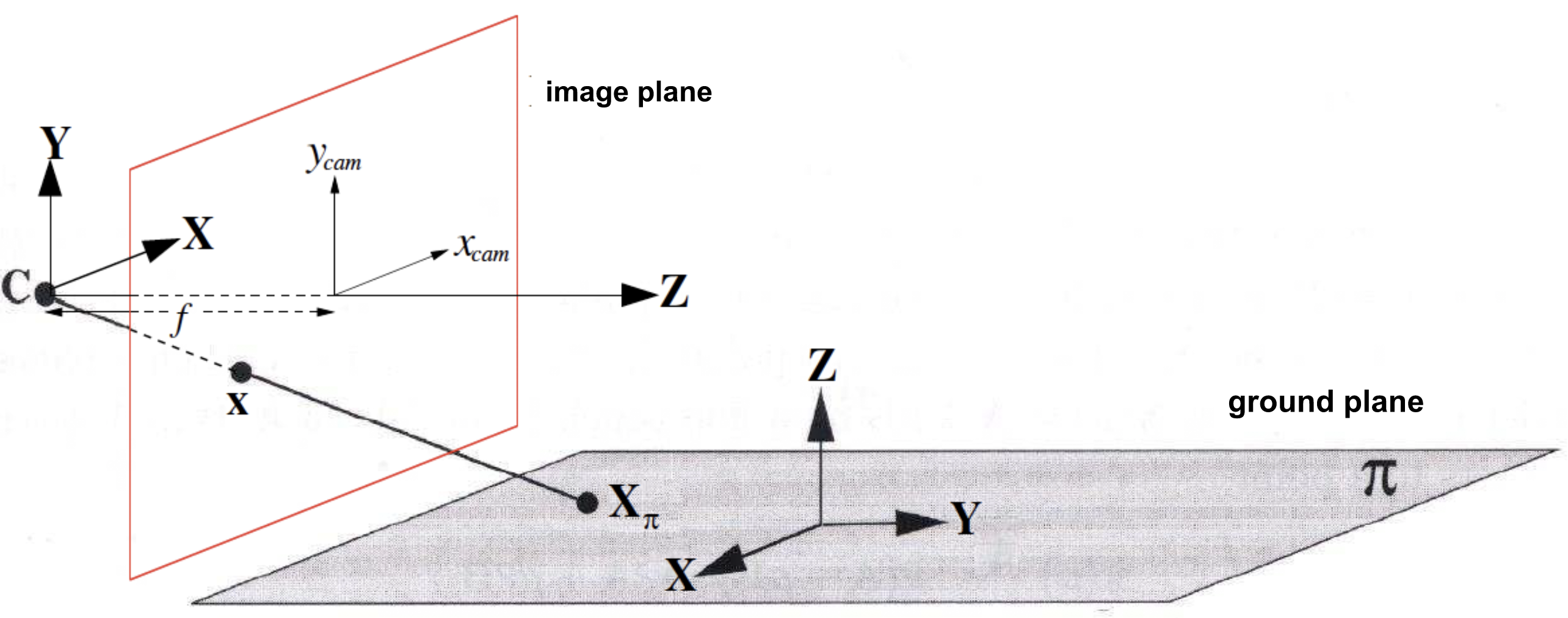}
    \caption{Projection scheme for the pinhole camera model. Given a $\mathbf{x}$ point in the image plane, $\mathbf{X_{\tau}}$ is its spatial point (homogeneous coordinate) in the $\tau$ ground plane. Image taken and adapted from \cite{hartley_zisserman_2004}.}
  \label{fig:map_2d_3d}
\end{figure}

In this section, we describe two transformations which  compute a corresponding 3D position $\mathbf{\tilde{X}_{\pi}}$ (inhomogeneous) on a ground plane $\tau$ for each point $\mathbf{x}$ in the image coordinate and the world coordinate $\mathbf{\tilde{C}}$ of the camera centre $\mathbf{C}$ for the pinhole camera model (Figure \ref{fig:map_2d_3d}).

We denote by $\mathbf{X}$ a point in space with world coordinates represented by a homogeneous vector $(\mathrm{X},\mathrm{Y},\mathrm{Z},1)^{T} \in \mathbb{R}^{4}$.
Its homogeneous coordinates on the image plane $\mathbf{x}\in \mathbb{R}^{3}$ can be obtained by:
\begin{equation}
    \mathbf{\mathbf{x} = \mathbf{PX}}
    \label{eq:mapping_equation}
\end{equation}
where $\mathbf{P}$ is the $3\times 4$ homogeneous \textit{camera projection matrix} written as:
\begin{equation}
    \mathbf{P = K [ R | t ]},
    \label{eq:general_form_2}
\end{equation}
with $\mathbf{K}$ being the $3\times3$ \textit{intrinsic matrix} and $\mathbf{[ R | t ]}$ being the $3\times4$ \textit{extrinsic matrix} where $\mathbf{R}$ is  the $3\times3$ rotation matrix and $\mathbf{t}$ is the $3\times1$ translation vector. 

Let $\mathbf{M = KR}$ and $\mathbf{p_4}$ is the last column vector of $\mathbf{P}$, the matrix $\mathbf{P}$ then can be written in the form
\begin{equation}
    \mathbf{P = [ M | p_4 ]}.\,
    \label{eq:M_represent}
\end{equation}
Since $\mathbf{K}$ and $\mathbf{R}$ in the finite projective camera model \cite{hartley_zisserman_2004} are assumed to be non-singular, this means that $\mathbf{M}$ is also  non-singular. We then can compute $\mathbf{\tilde{C}}$ by: 
\begin{equation}
    \mathbf{\tilde{C} = - M^{-1}p_4}
    \label{eq:camera_3D}
\end{equation}

To specify $\mathbf{X_{\pi}}$ on the ground plane $\pi$, we need further constraints \cite{hartley_zisserman_2004,sagererentwicklung}. In this setting, the coordinate systems of the camera and the surrounding space both have the $z$-axis as the height axis and the back projection line of an image point $\mathbf{x}$ hits the ground exactly when the $z$-coordinate of this line is set to zero since the origin of the spatial coordinate system is lying on the ground. The inhomogeneous coordinate $\mathbf{\tilde{X}}_{\pi} \in \mathbb{R}^{3}$ of $\mathbf{X}_{\pi}$ given a point $\mathbf{x}$ in the image coorindate is estimated by:
\begin{equation}
    \mathbf{\tilde{X}_{\pi} = - } \frac{\tilde{C_3}}{\tilde{X_3}} \mathbf{M^{-1}x + \tilde{C}}
    \label{eq:final_project}
\end{equation}
where $\tilde{C_3}$ and $\tilde{X_3}$ are obtained by:
\begin{equation}
    \begin{split}
[\,\tilde{X_1}, \tilde{X_2}, \tilde{X_3} ]\,^{T} &= \mathbf{M^{-1}x}  \\
[\, \tilde{C_1}, \tilde{C_2}, \tilde{C_3} ]\,^{T} &= \mathbf{\tilde{C}.}
\end{split}
\end{equation}

\section{Pre-Clustering Algorithm}

\begin{figure}[!hbt]
\centering
\includegraphics[width=0.5\textwidth]{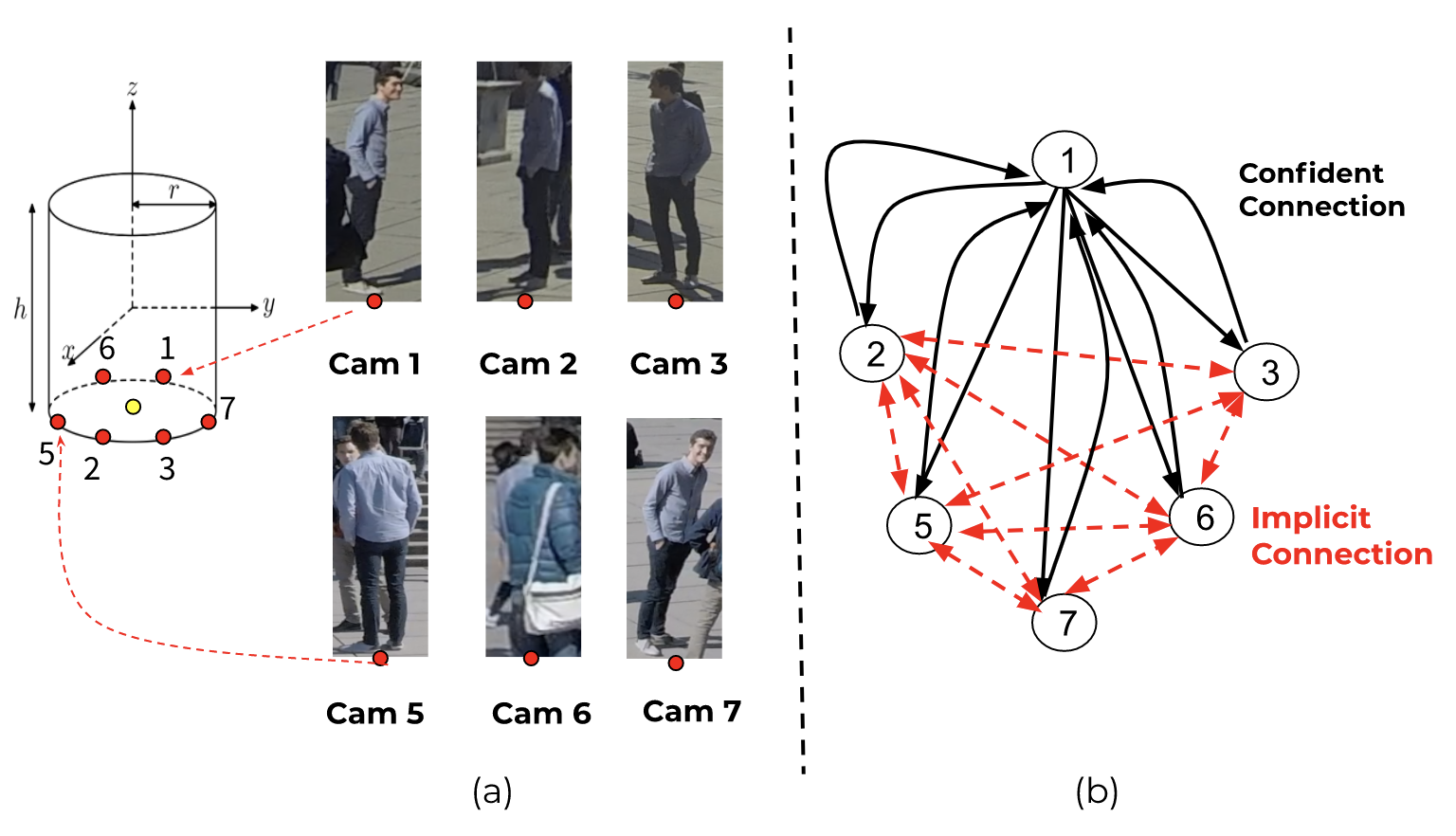}
\caption{(a) The geometry constraint of 2D foot points belong to the same object on the 3D space, (b) An illustration of our pre-clustering in WILDTRACK with 7 cameras where the object is not visible at camera 4.}
\label{fig:2Dclustering}
\end{figure}

We present in Algorithm \ref{alg:pre-clustering} the 3D geometry projection-based pre-clustering whose input is a set of detections across cameras at each time frame $t$ and the output is a set of clusters for each object appearing in the scene. The linear assignment problem at line $8$ exploits the 3D geometry constraints among points in the same object depicted in Figure \ref{fig:2Dclustering}-a.  The code line 13 in algorithm \ref{alg:pre-clustering} will verify in reverse directions (confident connections) for nodes in an initial cluster $I_b$ to alleviate errors attained in inaccurately calibrated cameras and noisy detections. Figure \ref{fig:2Dclustering}-b illustrates this procedure with $I_{1} = \{2,\, 3,\, 5,\, 6,\, 7\}$. 

\scalebox{0.8}{
\begin{minipage}{1.2\linewidth}
\removelatexerror
    \begin{algorithm}[H]
        \SetAlgoLined
        \DontPrintSemicolon
        \KwInput{
        Timeframe $t$, detections $B$, $r$ is a radius to scan nearby detections.
        }
        \KwOutput{$C = \{C_b: b \in B,\,\dettime(b) = t\}$ is a set of clusters for each detection at timeframe $t$.
        }
        $D^{t} \leftarrow \{b \in B: \dettime(b) = t\}$ \\
        $C \leftarrow \emptyset $ \\ 
        \tcp{Generate matches for each object}
        \For{$b \in D^{t}$}{
        \tcp{Initialize}
        $j \leftarrow \cam(b)$, $I_b \leftarrow \emptyset$,  $C_b \leftarrow \emptyset$\\
        \tcp{{\fontsize{9.5}{9.5}\selectfont Define matching candidates in cam $j$}} $B^{t,j}(b) = \{b' \in D^{t}: \mathrm{dist}(h(b), h(b')) \leq r, \cam(b') = j  \}$\\
        \tcp{Check one way connection}
        \For{camera $j' \neq j$}{
        \tcp{{\fontsize{9.5}{9.5}\selectfont Define matching candidates in cam $j'$}} $B^{t,j'}(b) = \{b' \in D^{t}: \mathrm{dist}(h(b), h(b')) \leq r,  \cam(b') = j' \}$\\
        Solve linear assignment problem between $B^{t,j}(b)$ and $B^{t,j'}(b)$ with costs $\mathrm{dist}(h(b_1),h(b_2))$ for $b_1 \in B^{t,j}(b)$, $b_2 \in B^{t,j'}(b)$\;
        \If{$b$ is matched to some node $b' \in B^{t,j'}(b)$}{
        $I_b \leftarrow I_b \cup \{b'\}$\\ \label{alg:update_math}
        }
        }
        \tcp{Clustered points must be matched both ways (confident connections)}
        $C_b = \{b' \in I_b: b \in I_{b'}\}$\\
        \tcp{Update set of clusters}
        $C \leftarrow C \cup \{C_b\}$
        }
        \textbf{return} $C$
        \caption{Pre-Clustering}
        \label{alg:pre-clustering}
    \end{algorithm}\end{minipage}}

\section{Learning Spatial-Temporal Affinity and Correcting ID-Switch Positions}
\subsection{Sampling Training Data}
We describe in Algorithm \ref{alg:samplimg-spatial}, \ref{alg:samplimg-temporal} the training data sampling for $f_{\mathrm{spatial}},\, f_{\mathrm{temporal}}$ networks used in generating affinity costs (Eqs. \ref{eq:temporal-cost},\,\ref{eq:spatial-cost}), where $\mathrm{Uni(a,b)}$ is the uniform distribution between $a$ and $b$, $\mathrm{id}(x)$ is the single ID of tracklet $x$. $\mathrm{Pos}_{s}, \, \mathrm{Neg}_{s}$ are sets of samples in the same and different objects connected by spatial edges. 
Likely, $\mathrm{Pos}_{t}, \, \mathrm{Neg}_{t}$ are outputs for temporal edges. For $f_{split}$, we create the training data by running the trained CenterTrack on the validation set in each dataset.

\subsection{Network Architecture}
\label{appendix:network_architecture}
\begin{figure}[!hbt]
\centering
\includegraphics[width=0.4\textwidth]{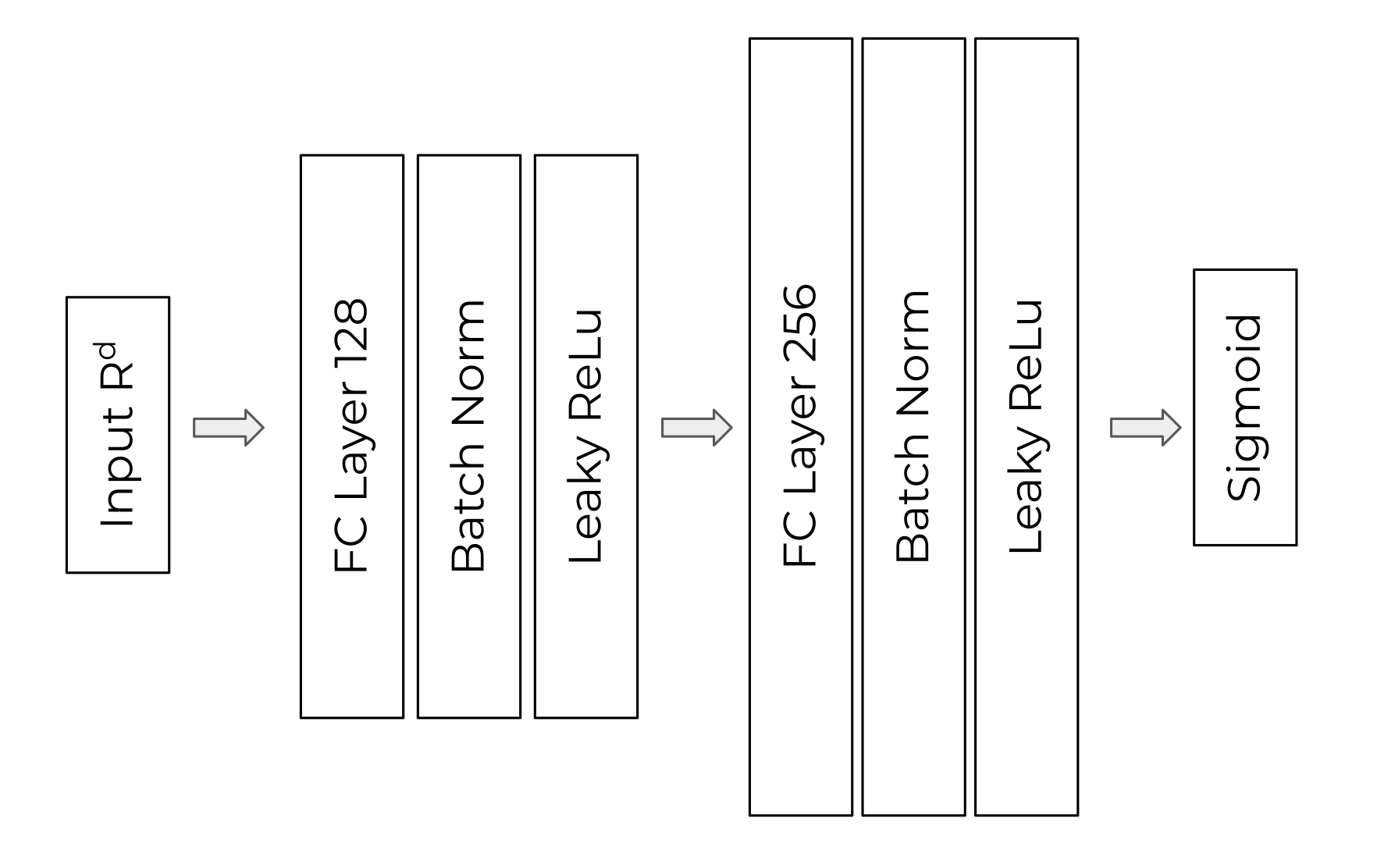}
\caption{The binary network architecture used to train for $f_{\mathrm{spatial}},\, f_{\mathrm{temporal}}$ and $f_{\mathrm{split}}$.}
\label{fig:network}
\end{figure}

For both $f_{\mathrm{spatial}},\, f_{\mathrm{temporal}}$ and $f_{\mathrm{split}}$, we employed the same architecture as depicted in Figure \ref{fig:network} given the input feature vectors with $d$ dimensions. We train $f_{\mathrm{spatial}},\, f_{\mathrm{temporal}}$ with $2$ epochs and $f_{\mathrm{split}}$ with $30$ epochs using Adam optimizer \cite{kingma2014adam}. A modified F1 loss function~\cite{johnson2019survey} is also applied during the training process to mitigate problems caused by imbalanced training data.

\section{Solver for Lifted Multicut for Tracking}
\label{sec:solver-lmp}
Since the lifted multicut problem is NP-hard to solve~\cite{keuper2015lifted}, we resort to efficient heuristics.
We first compute a preliminary partition with the Greedy Additive Edge Contraction (GAEC) algorithm, followed by an improvement step with the Kernighan-Lin local search (KL-Local) procedure \cite{keuper2015lifted}. We summarize the used solver in an Algorithm \ref{alg:solver-lifted}.
These heuristics have been shown to yield high quality solutions in practice~\cite{levinkov2017comparative}.

\vspace{0.2in}
\scalebox{0.9}{
\begin{minipage}{1.0\linewidth}
\removelatexerror
\begin{algorithm}[H]
	\SetAlgoLined
	\DontPrintSemicolon
	\KwInput{Spatial-temporal graph
	$G = (V, F),\, F = E^{t} \cup E^{s} \cup E^{c}$ \\
	\hspace{0.42in} Edge costs $c_{e^{t}},\, c_{e^{s}},\, M$ 
	}
	\KwOutput{\vspace{-0.1in}
	Edge label function $y: F \rightarrow \{0,\,1\}$
	}
	\vspace{0.1in}
	$y \leftarrow \mathrm{Run\ GAEC}\,(G,\,c_{e^{t}},\, c_{e^{s}},\, M)$\\
	$y \leftarrow \mathrm{Run\ KL-Local}\,(G,\,c_{e^{t}},\, c_{e^{s}},\, M,\,y)$\\
	\textbf{return} $y$
	\caption{Solver for LMGP Tracker}
	\label{alg:solver-lifted}
\end{algorithm}\end{minipage}} 

\section{Generating Tracking Predictions in 3D Space}
\label{sec:3d-interpolation}
Given clusters of the spatial-temporal tracking graph, we interpolate trajectories in 3D-coordinates.
Since a cluster consists of a set of tracklets, which in turn consist of a set of detections, we directly associate a cluster with its underlying set of detections.
Hence, let a cluster consisting of detections $\{b_1,\ldots,b_s\}$ be given.
At each timestep $t$ such that there exists at least one detection in that timestep, we obtain a 3D position through
\begin{equation}
\begin{split}
p_{avg}^t & = \frac{\sum_{i : \dettime(b_i) = t} h(b_i)}{\abs{\{i : \dettime(b_i) = t\}}} \\
p_{3D}^t & = \argmax_{p : \norm{p - p_{avg}^{t}} \leq r } \sum_{i : \dettime(b_i) = t} \norm{\pi_{\cam(b_i)}(p), b_i}_{iou}^2
\end{split}
\label{eq:3d-interpolation}
\vspace{-0.5in}
\end{equation}
where $\pi_j(p)$ is the projection of a 3D cylinder centering at $p$ to a rectangle bounding box on 2D coordinates of camera $j$.
The above interpolation step in Equation \ref{eq:3d-interpolation} first takes the average position of the 3D-projections of all relevant detections of a given timepoint.
Second, a 3D position in the vicinity of the mean position is calculated such that the reprojection error w.r.t. IoU is minimized.

\section{Experimental Results}
\subsection{Implementation Details}
\label{sec:implement}
\myparagraph{Generating Tracklets}
For both PETS-09 \cite{ellis2009pets2009}, Campus \cite{xu2016multi}, and WILDTRACK \cite{chavdarova2018wildtrack} datasets, we use CenterTrack~\cite{zhou2020tracking} trained on the CrowdHuman dataset \cite{shao2018crowdhuman} and pre-trained it on training sequences in each dataset with the following settings: heatmap noise $\, 0.05\, $, tracklets confidence $0.4$, false positive rate $0.1$, and batch size of $32$ images. All other parameters are identical with the default settings. Given trained models, tracklets are generated by running CenterTrack with the provided detections and employ a tracking threshold $0.2$, pre-threshold $0.4$, which are scores for predicting a bounding box and feeding the heatmap to the next frame, respectively.

\scalebox{0.8}{
\begin{minipage}{1.2\linewidth}
\removelatexerror
\begin{algorithm}[H]
	\SetAlgoLined
	\SetKwFor{ForEach}{for each}{do}{end}
	\DontPrintSemicolon
	\KwInput{
		$\Gamma = \{\tau: \mathrm{cam}(\tau) \in \{1,2,\ldots,m \} \}$ is a set of all ground truth trajectories across $m$ cameras. \\
		\hspace{0.4in} $n$ is the number of random sampling.
	}
	\KwOutput{\vspace{-0.1in}
		\begin{equation*}
			\hspace{-0.5in}
			\begin{split}
				\mathrm{Pos}_{s} = \{(x,y): \mathrm{cam}(x) \neq \mathrm{cam}(y),  \mathrm{id}(x) = \mathrm{id}(y),\, \\ \mathrm{time}(x) \cap \ \mathrm{time}(y) \neq \varnothing\}\\
				\mathrm{Neg}_{s} = \{(x,y): \mathrm{cam}(x) \neq \mathrm{cam}(y),  \mathrm{id}(x) \neq  \mathrm{id}(y), \,\\ \mathrm{time}(x) \cap \mathrm{time}(y) \neq \varnothing\}
			\end{split}
	\end{equation*}}
	\vspace{-0.1in}
	\tcp{Initialize}
	$\mathrm{Pos}_{s} \leftarrow \emptyset,\, \mathrm{Neg}_{s} \leftarrow \emptyset$
	\\
	\Repeat{n times}{
		\ForEach{camera $j \in \{1,\ldots ,m \}$}{
			\ForEach{$\tau = (b_1, b_2,\ldots, b_{|\tau|}) : \mathrm{cam}(\tau) = j$}{
				\tcp{Randomly remove an interval in $\tau$} 
				$i \sim \textrm{Uni}(1, |\tau|),\, k \sim \textrm{Uni}(i, |\tau| - i - 1) $\\
				$x \leftarrow (b_1,\ldots,b_{i-1}),\  y  \leftarrow (b_{i+k+1}),\ldots, b_{|\tau|})$\\
				$\mathrm{R}_{\tau}^{j} \leftarrow \{x,y\}$\\
			}
		}
		
		\tcp{Initialize balanced sampling}
        $n_{\textrm{balance}} \leftarrow  0 $ \\ 
		\tcp{Generate positive pairs}
		    \ForEach{$\tau, \tau' : \mathrm{cam}(\tau) \neq \mathrm{cam}(\tau') \textbf{ and } \mathrm{id}(\tau) = \mathrm{id}(\tau') \textbf{ and } \dettime(\tau) \cap  \dettime(\tau') \neq  \emptyset$}{
		        \ForEach{$(x,y) \in \mathrm{R}_{\tau}^{\mathrm{cam}(\tau)} \times \mathrm{R}_{\tau'}^{\mathrm{cam}(\tau')}$}{
		        $\mathrm{Pos}_{s} \leftarrow \mathrm{Pos}_{s} \cup \{(x,y)\}$\\
	        	$n_{\textrm{balance}} \leftarrow  n_{\textrm{balance}}+1$
		        
		        }
		    }
	    \tcp{Generate negative pairs}
	    \ForEach{$\tau, \tau' : \mathrm{cam}(\tau) \neq \mathrm{cam}(\tau') \textbf{ and } \mathrm{id}(\tau) \neq \mathrm{id}(\tau') \textbf{ and } \dettime(\tau) \cap  \dettime(\tau') \neq \emptyset$}{
		        \ForEach{$(x,y) \in \mathrm{R}_{\tau}^{\mathrm{cam}(\tau)} \times \mathrm{R}_{\tau'}^{\mathrm{cam}(\tau')}$}{
		        \tcp{Balanced Sampling}
		        \If{$n_{\textrm{balance}} > 0 $}{
		        $\mathrm{Neg}_{s} \leftarrow \mathrm{Neg}_{s} \cup \{(x,y)\}$\\
		        	$n_{\textrm{balance}} \leftarrow  n_{\textrm{balance}}-1$
	        	}
		        }
		    }
	}
	\textbf{return} $\mathrm{Pos}_{s},\, \mathrm{Neg}_{s}$
	\caption{Sampling spatial edges}
	\label{alg:samplimg-spatial}
\end{algorithm}\end{minipage}}
\myparagraph{Multi-view Appearance Feature}
We utilize DG-Net \cite{zheng2019joint}, one of the best performing methods for re-identification to extract embedding vectors for detections.
To train DG-Net, the training data is selected from pre-clustering of unoccluded detections $C'_{b}$ for each object (Section 3.1, main paper). We use the best models trained on Market-151~\cite{zheng2015scalable} and DukeMTMC-Reid~\cite{ristani2016performance} as the ``teacher network'' in DG-Net. Noting that pre-training on other datasets is a common practice in MOT, and pre-training on the above datasets was e.g.\ done in \cite{hornakova2020lifted}. While our method benefits from strong appearance features from DG-Net, these features alone are not mainly responsible for our performance, see Table 1 and 2 in the main paper with \texttt{LMGP w/o Pre-Clustering}, which mainly relies on appearance features and does not include 3D-information.

\scalebox{0.8}{
\removelatexerror
\begin{minipage}{1.2\linewidth}
\begin{algorithm}[H]
	\SetAlgoLined
	\DontPrintSemicolon
	\SetKwFor{ForEach}{for each}{do}{end}
	\KwInput{
		$\Gamma = \{\tau: \mathrm{cam}(\tau) \in \{1,2,\ldots,m \}\}$ is a set of all trajectories across $m$ cameras. \\
		\hspace{0.4in} $n$ is the number of random sampling.
	}
	\KwOutput{\vspace{-0.1in}
		\begin{equation*}
			\hspace{-0.5in}
			\begin{split}
				\mathrm{Pos}_{t} = \{(x,y): \mathrm{cam}(x) = \mathrm{cam}(y),  \mathrm{id}(x) = \mathrm{id}(y),\, \\ \mathrm{time}(x) \cap \ \mathrm{time}(y) = \varnothing\}\\
				\mathrm{Neg}_{t} = \{(x,y): \mathrm{cam}(x) = \mathrm{cam}(y),  \mathrm{id}(x) \neq  \mathrm{id}(y), \,\\ \mathrm{time}(x) \cap \mathrm{time}(y) = \varnothing\}
			\end{split}
	\end{equation*}}
	\tcp{Initialize}
	$\mathrm{Pos}_{t} \leftarrow \emptyset,\, \mathrm{Neg}_{t} \leftarrow \emptyset$\\
	\Repeat{n times}{
		\textrm{Select} randomly camera $j$\\
		\tcp{Initialize balanced sampling}
        $n_{\textrm{balance}} \leftarrow  0 $ \\ 
		\ForEach{$\tau = (b_1, b_2,\ldots, b_{|\tau|}) :  \mathrm{cam}(\tau) = j$}{
			\tcp{Randomly remove an  interval in $\tau$}  
			$i \sim \textrm{Uni}(1, |\tau|),\, k \sim \textrm{Uni}(i, |\tau| - i - 1)$\\
            $x \leftarrow (b_1,\ldots,b_{i-1}),\  y \leftarrow (b_{i+k+1},\ldots, b_{|\tau|})$\\
			\tcp{Generate positive pairs}
			$\mathrm{Pos}_{t} \leftarrow \mathrm{Pos}_{t} \cup \{(x,y) \}$ \\
			$R_{\tau} \leftarrow \{x,y\}$ \\ 
			$n_{\textrm{balance}} \leftarrow  n_{\textrm{balance}}+1$ 
		}
 		\tcp{Generate negative pairs}

	    \ForEach{$\tau,\tau': \mathrm{id}(\tau) \neq \mathrm{id}(\tau')$ }{
	        \ForEach{$(x,y) \in R_{\tau} \times R_{\tau'}: \dettime(x) \cap \dettime(y) = \emptyset$}{
	        \tcp{Balanced Sampling}
	        \If{$n_{\textrm{balance}} > 0 $}{
	        $\mathrm{Neg}_{t} \leftarrow \mathrm{Neg}_{t} \cup \{(x,y)\}$ \\ 
	        $n_{\textrm{balance}} \leftarrow n_{\textrm{balance}} - 1$
	        }
	    }
	    }
}
	\textbf{return} $\mathrm{Pos}_{t},\, \mathrm{Neg}_{t}$
	\caption{Sampling temporal edges}
	\label{alg:samplimg-temporal}
\end{algorithm}\end{minipage}}

\subsection{Ablation Study of Affinity Costs}
\label{sec:computing-affinity}
Since the performance of a tracking system depends highly on the accuracy of local connections, this section validates the effect of several proposed affinities, which form our edge costs
$c_{e^{t}}$ and $c_{e^{s}}$. In particular,
\begin{equation}
    \begin{split}
c_{e^{t}} = f_{\mathrm{temporal}}(c_{\mathrm{index}}^{\mathrm{app}}(\tau,\tau'), \ c^{\mathrm{fw,t}}(\tau,\tau'),\  c^{\mathrm{bw,t}}(\tau,\tau'),\, \\
t(\tau, \tau'))
\end{split}
\label{eq:temporal-cost}
\end{equation}

\vspace{-0.1in}
\begin{equation}
    \begin{split}
    c_{e^{s}} = f_{\mathrm{spatial}}(c^{\mathrm{fw,s}}(\tau,\tau'),\  c^{\mathrm{bw,s}}(\tau,\tau'),\ c^{\mathrm{app}}(\tau,\tau'),\ \\ c^{\mathrm{avg 3D}}(\tau,\tau'),  c^{\mathrm{pc}}(\tau,\tau')) 
\end{split}
\label{eq:spatial-cost}
\end{equation}

where $\mathrm{index} \in \{\mathrm{best,min,max,avg,std}\}$, described at the Equation (4), Subsection 3.2 in the main paper. For spatial cost in Equation \ref{eq:spatial-cost}, the appearance part $c^{\mathrm{app}}(\tau,\tau')$ only uses the visible detections at current cameras.


Table \ref{tab:temporal-costs-performance} and \ref{tab:spatial-costs-performance} present ablation studies on WILDTRACK dataset for different settings where positive/negative refers to edge pairs whose nodes are in the same/distinct pedestrians.

In Table \ref{tab:temporal-costs-performance}, we compare our full settings for temporal edge costs $c_{e^{t}}$ in Equation \ref{eq:temporal-cost}  and compare with: (i) without using forward/backward prediction (second row) and (ii) without utilizing pre-clustering results to query images at other cameras (third row), i.e., we only measure appearance affinities for $\tau$ and $\tau'$ in the current camera. It can be observed that pre-clustering is the most critical factor; thereby, the total accuracy will decline from $99\%$ down to $94\%$ in the Acc metric. Secondly, the forward/backward affinities also contribute significantly to the final performance with a $2\%$ improvement. 

For spatial edges in Table \ref{tab:spatial-costs-performance} we examine the impact of two ingredients: (i) forward/backward prediction (second row) and (ii) without applying our novel prior-3D based on pre-clustering agreement affinity (third row). The obtained results confirm that prior-3D affinity is the most influential affinity that is able to boost edge costs accuracy up to $7\%$, yielding $100\%$ accuracy for both positive and negative edges. Similarly as for spatial edges, the forward/backward affinities also contribute to complement our prediction with a growth of approximately $2\%$.
\begin{table}[!hbt]
  \centering
  \scalebox{0.7}{
  \begin{tabular}{c|c|c|c|c|c}
\Xhline{2\arrayrulewidth}
    \multicolumn{1}{c|}{Method} & \multicolumn{1}{c|}{Type } & \multicolumn{1}{c|}{Precision $\%$} & \multicolumn{1}{c|}{Recall $\%$} & \multicolumn{1}{c|}{F1 $\%$} & \multicolumn{1}{c}{Acc $\%$}\\ \Xhline{2\arrayrulewidth}
     \multirow{2}{*}{\STAB{\rotatebox[origin=c]{0}{Full settings}}}
     & positive                      &   98                        &  90                         &  94                         &        \multirow{2}{*}{\STAB{\rotatebox[origin=c]{0}{99}}}                     \\
     & negative                      &     99                      &  100                         &  100                                             \\
 \hline
          \multirow{2}{*}{\STAB{\rotatebox[origin=c]{0}{W/o Forward, backward}}}
     & positive                      &   95                        &  89                         &  92                         &  \multirow{2}{*}{\STAB{\rotatebox[origin=c]{0}{97}}}                      \\
     & negative                      &  97                        &  98                         &  98                                             \\
\hline
     \multirow{2}{*}{\STAB{\rotatebox[origin=c]{0}{W/o Pre-clustering}}}
     & positive                      &   94                        &  88                         &  91                         &  \multirow{2}{*}{\STAB{\rotatebox[origin=c]{0}{94}}}                       \\
     & negative                      &     95                      &  94                         &  95                                                \\
\Xhline{2\arrayrulewidth}
  \end{tabular}}
  \vspace{0.05in}
  \caption{Temporal edge costs performance with variant settings over $50565$ samples in which $3678$ edges are positive and the remaining of $46887$ are negative.}
  \label{tab:temporal-costs-performance}
\end{table} 

\begin{table}[!hbt]
  \centering
  \scalebox{0.7}{
  \begin{tabular}{c|c|c|c|c|c}
\Xhline{2\arrayrulewidth}
    \multicolumn{1}{c|}{Method} & \multicolumn{1}{c|}{Type } & \multicolumn{1}{c|}{Precision $\%$} & \multicolumn{1}{c|}{Recall $\%$} & \multicolumn{1}{c|}{F1 $\%$} & \multicolumn{1}{c}{Acc $\%$}\\ \Xhline{2\arrayrulewidth}
     \multirow{2}{*}{\STAB{\rotatebox[origin=c]{0}{Full settings}}}
     & positive                      &   100                        &  100                         &  100                         &        \multirow{2}{*}{\STAB{\rotatebox[origin=c]{0}{100}}}                     \\
     & negative                      &     100                      &  100                         &  100                                             \\
 \hline
          \multirow{2}{*}{\STAB{\rotatebox[origin=c]{0}{W/o Forward, backward}}}
     & positive                      &   97                        &  99                         &  98                         &  \multirow{2}{*}{\STAB{\rotatebox[origin=c]{0}{98}}}                      \\
     & negative                      &  98                        &  99                         &  98                                             \\
\hline
     \multirow{2}{*}{\STAB{\rotatebox[origin=c]{0}{W/o Prior-3D}}}
     & positive                      &   92                        &  94                         &  93                         &  \multirow{2}{*}{\STAB{\rotatebox[origin=c]{0}{93}}}                       \\
     & negative                      &     94                      &  93                         &  94                                                \\
\Xhline{2\arrayrulewidth}
  \end{tabular}}
  \vspace{0.05in}
  \caption{Spatial edge costs performance with different configurations evaluated over $11026$ samples with $3215$ edges are positive and $7811$ remaining are negative.}
  \label{tab:spatial-costs-performance}
\end{table} 

\begin{figure*}[!hbt]
\centering
\includegraphics[width=1.0\textwidth]{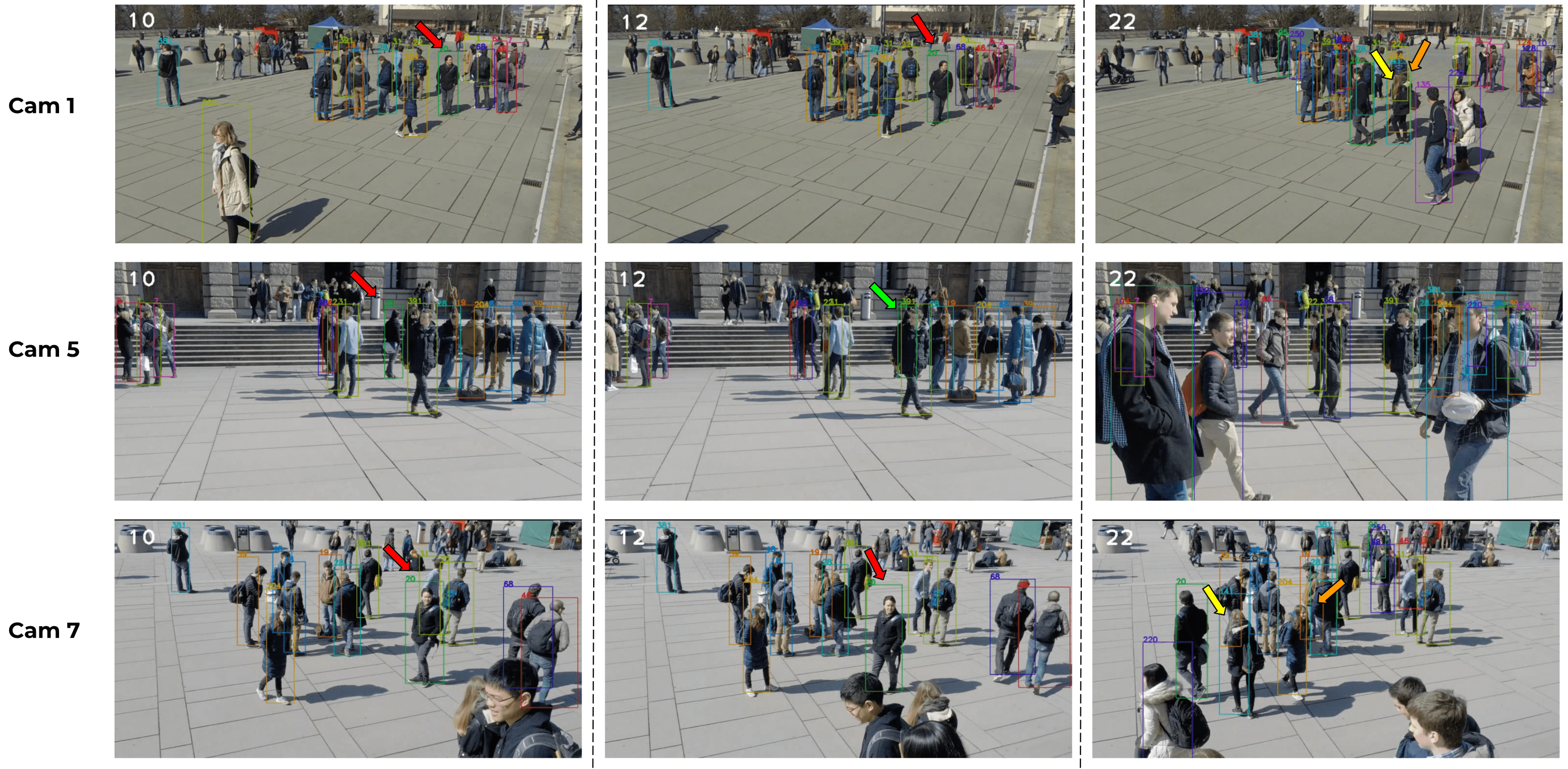}
\caption{Qualitative tracking results on WILDTRACK at three overlapping camera views (zoom in for better visualization). The superiority of multi-cameras can be seen in the two following scenarios. First, even an object $20$ is fully occluded at frame $12$ in Cam $5$ (green arrow), our tracker could still track this target by establishing inter-camera correspondences in Cam $1$ and $7$ (red arrow). Second, at frame $22$ in Cam $1$, the ID-Switch error will occur between ID $141$ and $204$ (yellow and orange arrows) due to ambiguous appearance affinities; fortunately, we could avoid this error by harnessing corresponding detections for those objects in Cam $7$, which are separable.}
\label{fig:qualitative-results}
\end{figure*}
\begin{figure*}[!t]
\centering
\includegraphics[width=1.0\textwidth]{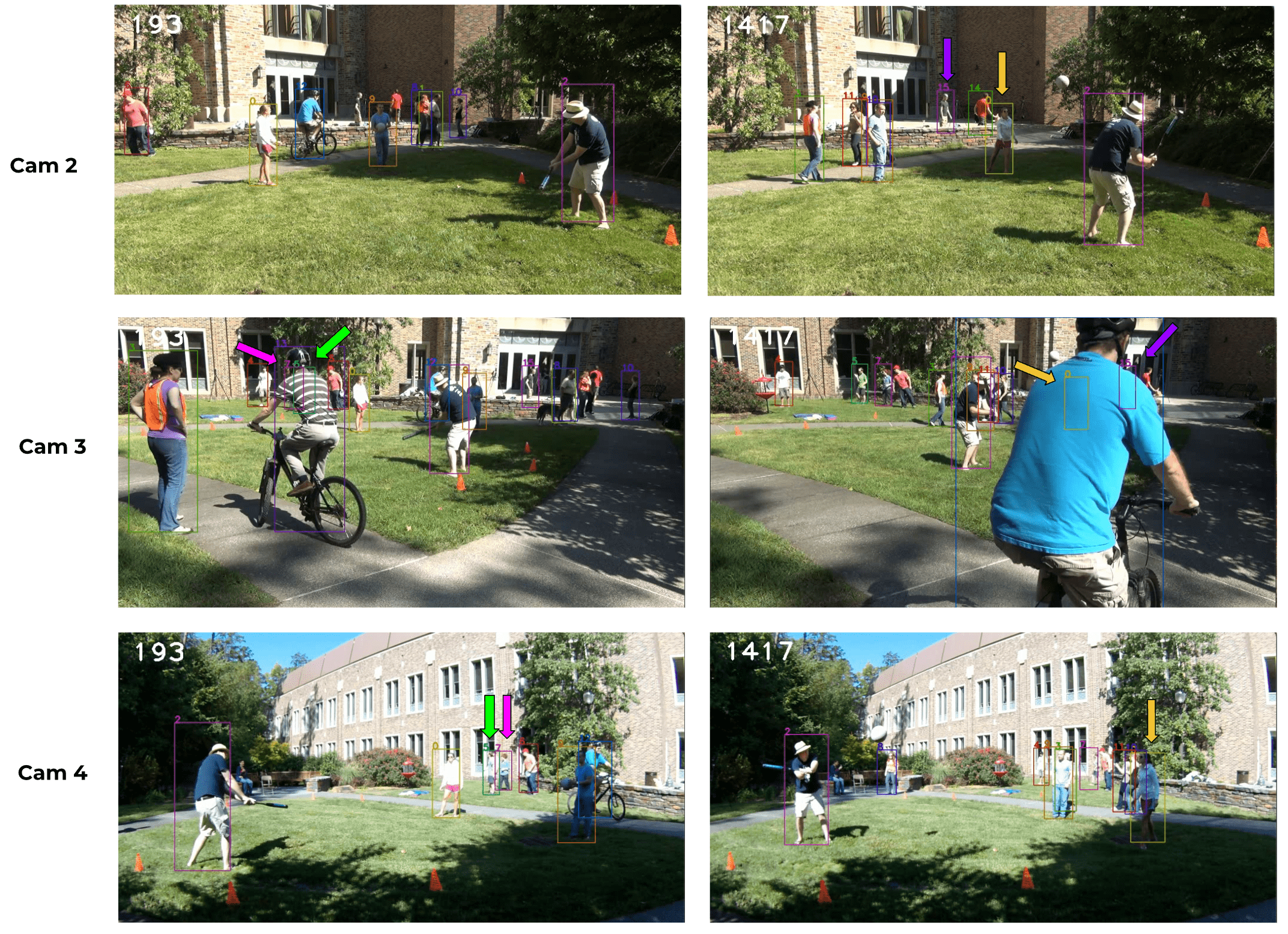}
\caption{Qualitative tracking results in the Garden 1 in Campus dataset at three overlapping camera views (zoom in for better visualization). The benefit of the multi-camera approach can be found in two typical situations. First, at frame 193 (left column), two objects at the pink and green arrows in Cam 3 were not visible in this view; however, we could overcome these missing positions by using information from Cam 4. Similarly, at frame 1417, two objects at yellow and purple arrows are occluded in Cam 3. Fortunately, leveraging visible detections in Cam 4 and Cam 2, we can continue tracking these objects.}
\label{fig:qualitative-results-campus}
\end{figure*}
\subsection{Performance of Splitting Tracklets in the Spatial-Temporal Tracking Graph}

We now demonstrate that the pre-clustering is helpful during the construction of the spatial-temporal tracking graph.
In particular, we experimented on the WILDTRACK dataset due to its severe occlusions with two current state of the art single-camera pedestrian trackers: CenterTrack~\cite{zhou2020tracking} and Tracktor++~\cite{bergmann2019tracking}.
Given tracklets computed by these trackers, we go through each tracklet and assign "correct" label for a pair of two consecutive detections if they are of the same object and as an "ID-Switch" label otherwise.
Table~\ref{tab:center-tracktor} shows our performance. The results indicate that we can detect and correct up to $97\%$ of ID-Switches w.r.t.\ the F1 score for both CenterTrack and Tracktor++ over a total of 310 resp.\ 280 tracklets on the testing frames.
Furthermore, we were also able to retain correct consecutive detection for both trackers ($100 \%$ F1 score).
Obtaining such high-quality tracklets not only allows us to reduce ID-Switch errors significantly but also allows for better affinity computations, resulting in an overall better final tracking result. 

\begin{table}[!hbt]
  \centering
  \scalebox{0.7}{
  \begin{tabular}{c|c|c|c|c|c}
\Xhline{2\arrayrulewidth}
    \multicolumn{1}{c|}{Method} & \multicolumn{1}{c|}{Type } & \multicolumn{1}{c|}{Precision $\%$} & \multicolumn{1}{c|}{Recall $\%$} & \multicolumn{1}{c|}{F1 $\%$} & \multicolumn{1}{c}{$\#$Samples}\\ \Xhline{2\arrayrulewidth}
     \multirow{2}{*}{\STAB{\rotatebox[origin=c]{0}{CenterTrack 
     \cite{zhou2020tracking}}}}
     & Correct                      &   100                        &  100                         &  100                         &  3268                      \\
     & ID-Switch                      &     96                      &  98                         &  97                        &  310                        \\
 \hline
          \multirow{2}{*}{\STAB{\rotatebox[origin=c]{0}{Tracktor++ \cite{bergmann2019tracking}}}}
     & Correct                      &   100                        &  100                         &  100                         &  3256                     \\
     & ID-Switch                      &  97                        &  96                         &  97                         &  280                      \\
\hline
\Xhline{2\arrayrulewidth}
  \end{tabular}}
  \caption{Our performance in reducing ID-switch errors and retaining correct detection pairs in the tracklet splitting step. The result is measured with two state-of-the-art single-camera trackers.}
  \label{tab:center-tracktor}
\end{table} 
\subsection{Single-Camera Benchmark}
\begin{table}[!hbt]
\scalebox{0.6}{
\begin{tabular}{c|c|c|c|c|c|c|c}
\Xhline{2\arrayrulewidth}
Method& IDF1 $\uparrow$ & MOTA  $\uparrow$ & MT  $\uparrow$   & ML  $\downarrow$  & FP $\downarrow$ & FN $\downarrow$ & IDs $\downarrow$ \\ \Xhline{2\arrayrulewidth}
Deep MCD + ptrack \cite{chavdarova2018wildtrack} & 64.3          & 52.4                              & 83.4         & 10.7         & 1023        & 2239        & 711          \\ 
Tracktor++   \cite{bergmann2019tracking}         & 65.0          & 60.4                               & 68.3          & 8.7        & 819         & 637        & 272          \\
CenterTrack   \cite{zhou2020tracking}    & 67.4         & 67.5                               & 74.6         & 3.4        & 649        & 494         & 278          \\ 
LMGP-single cam, detection       & 69.6         & 66.4                               & 75.3         & 5.2        & 690        & 575         & 128          \\ 
LMGP-single cam, tracklets       & 71.2        & 72.8                               & 82.5         & 2.7        & 755        & 394         & 154          \\ 
LMGP-multi-cam, tracklets             & \textbf{98.2} & \textbf{97.1}                      & \textbf{97.6} & \textbf{1.3} & \textbf{71} & \textbf{7}  & \textbf{12}  \\ \Xhline{2\arrayrulewidth}
\end{tabular}}
\caption{Our tracking performance compared to state of the art single camera methods on WILDTRACK.}
\label{tab:single-camera}
\end{table}

We provide experimental results on WILDTRACK (using the multi-camera setup and then projecting the result to a single camera, namely \texttt{LMGP-Multi-cam, tracklets}) compared to three modern single camera pedestrian tracking methods in Table \ref{tab:single-camera}. Furthermore, we also evaluate LMGP with a pure single-camera approach, including (i) using detections in the tracking graph (which reverts to the model of \cite{tang2017multiple}) and without pre-clustering or other 3D-geometry based components (\texttt{LMGP-single cam, detection}); (ii) using CenterTrack tracklets instead of detections and our full model (\texttt{LMGP-single cam, tracklets}).

As we mentioned, even with the best methods, the single-camera approach is still insufficient under severe occlusions.
Considering large improvements on all metrics, we argue that the multi-camera strategy convincingly improves upon the single-camera one in crowded settings.


\subsection{Running Time}
\begin{table}[!hbt]
\scalebox{0.7}{
\begin{tabular}{c|c|c|c|c}
\Xhline{2\arrayrulewidth}
Dataset& $\#$Camera   & $\#$Frame/Cam & Running Time & Speed (fps)\\ \Xhline{2\arrayrulewidth}
\small WILDTRACK & 7          & 400                             & 213         & 1.9               \\ 
\small PETS-09 S2-L1         & 3         & 795                              & 147          & 5.4              \\
\small PETS-09 S2-L2    & 3         & 436                              & 135         & 3.2                \\ 
\small PETS-09 S2-L3           & 3 & 240                      &111 & 2.2\\ \Xhline{2\arrayrulewidth}
\end{tabular}}
\caption{Running time and
speed of the LMGP tracker on
multi-camera sequences.}
\label{tab:running-time}
\end{table}

We show the running time of our tracker on two multi-camera datasets in Table \ref{tab:running-time}, which includes both feature extraction and data association linking steps. We implemented our tracker in $\textrm{C++}$ on a  Intel(R) Core(TM) i7-9800X CPU machine with $16$ cores and $126$ GB memory. Besides, four NVIDIA TITAN RTX GPUs with $24$GB memory are employed for parallel computation in appearance feature extractions and running the pre-clustering. In short, we conclude that our tracker runs reasonably fast; therefore, it is feasible to deploy it in real-world tracking applications.

\section{Qualitative Results}
Our qualitative tracking results are illustrated in Figure \ref{fig:qualitative-results} and Figure \ref{fig:qualitative-results-campus} for WILDTRACK and Campus, respectively, with three overlapping camera views. It is evident that monocular pedestrian tracking is insufficient to capture all objects due to highly crowded and cluttered scenes. For example, the target at the green arrow in Cam $5$ at frame $12$ in WILDTRACK is entirely occluded by other objects (Figure \ref{fig:qualitative-results}). In such situations, incorporating multiple views is a reasonable way to improve the total tracking performance.



\end{document}